%% file: tech_report.tex
\documentclass[british,english,american,11pt,a4paper]{article}
\usepackage[latin9]{inputenc}
\usepackage{color}
\usepackage{array}
\usepackage{verbatim}
\usepackage{units}
\usepackage{multirow}
\usepackage{amsmath}
\usepackage{amssymb}
\usepackage{graphicx}
\usepackage{setspace}
\usepackage[authoryear]{natbib}
\usepackage{xargs}[2008/03/08]

\makeatletter

\newcommand{\noun}[1]{\textsc{#1}}
\providecommand{\tabularnewline}{\\}

\usepackage{graphics,color}
\usepackage{natbib}
\setcitestyle{round}

\usepackage{wrapfig}
\usepackage{amssymb}
\usepackage{algorithm, algpseudocode}

\usepackage{url}

\usepackage{colortbl} 
\definecolor{headerColor}{rgb}{0.74,0.88,0.91}
\definecolor{yearPlanColor}{rgb}{0.85,0.93,0.95}

\makeatother

\usepackage{babel}
\begin{document}
\noindent \include{macros_hungv}

\noindent \include{coverpage}

\global\long\def\bhone{\bh^{1}}

\global\long\def\bhtwo{\bh^{2}}

\global\long\def\bhthree{\bh^{3}}

\global\long\def\bWone{\boldsymbol{W}^{1}}

\global\long\def\bWtwo{\boldsymbol{W}^{2}}

\global\long\def\bWthree{\boldsymbol{W}^{3}}

\global\long\def\bO{\boldsymbol{O}}

\global\long\def\be{\boldsymbol{e}}

\global\long\def\partfunc{\mathcal{Z}}

\global\long\def\lra#1{\left\langle #1\right\rangle }

\global\long\def\LL{\mathcal{L}}

\global\long\def\bW{\boldsymbol{W}}

\global\long\def\CDone{\text{CD}_{1}}

\global\long\def\CDm{\text{CD}_{m}}

\global\long\def\undernote#1#2{\underset{#2}{\underbrace{#1}}}

\global\long\def\model{\text{RBM}}

\pagebreak{}

\begin{doublespace}
\begin{center}
{\huge{}Energy-based Models for Video Anomaly Detection}
\par\end{center}{\huge \par}
\end{doublespace}

\begin{onehalfspace}
\begin{center}
\textbf{\large{}Hung Vu}{\large{}$\dagger$}\textbf{\large{}, Dinh
Phung}{\large{}$\dagger$}\textbf{\large{}, Tu Dinh Nguyen}{\large{}$\dagger$}\textbf{\large{},
Anthony Trevors}{\large{}$\ddagger$}\textbf{\large{} and}\\
\textbf{\large{}Svetha Venkatesh}{\large{}$\dagger$}\textbf{\large{}
}{\large{}}\\
{\large{}$\dagger$Centre for Pattern Recognition and Data Analytics
(PRaDA)}\\
{\large{}School of Information Technology, Deakin University, Geelong,
Australia}\\
{\large{}Email: \{hungv, tu.nguyen, svetha.venkatesh, dinh.phung\}@deakin.edu.au
}\\
{\large{}$\ddagger$Defence Science and Technology Organization (DSTO),
Melbourne, Australia}\\
{\large{}Email: Anthony.Travers@dsto.defence.gov.au}
\par\end{center}{\large \par}
\end{onehalfspace}
\begin{abstract}
Automated detection of abnormalities in data has been studied in research
area in recent years because of its diverse applications in practice
including video surveillance, industrial damage detection and network
intrusion detection. However, building an effective anomaly detection
system is a non-trivial task since it requires to tackle challenging
issues of the shortage of annotated data, inability of defining anomaly
objects explicitly and the expensive cost of feature engineering procedure.
Unlike existing appoaches which only partially solve these problems,
we develop a unique framework to cope the problems above simultaneously.
Instead of hanlding with ambiguous definition of anomaly objects,
we propose to work with regular patterns whose unlabeled data is abundant
and usually easy to collect in practice. This allows our system to
be trained completely in an unsupervised procedure and liberate us
from the need for costly data annotation. By learning generative model
that capture the normality distribution in data, we can isolate abnormal
data points that result in low normality scores (high abnormality
scores). Moreover, by leverage on the power of generative networks,
i.e. energy-based models, we are also able to learn the feature representation
automatically rather than replying on hand-crafted features that have
been dominating anomaly detection research over many decades. We demonstrate
our proposal on the specific application of video anomaly detection
and the experimental results indicate that our method performs better
than baselines and are comparable with state-of-the-art methods in
many benchmark video anomaly detection datasets. 
\end{abstract}

\section{Introduction\label{sec:intro}}

\input{intro.tex}

\section{Literature Review\label{sec:literature_review}}

\input{literature_review.tex}

\section{Energy-based Anomaly Detection}

\input{methodology.tex}

\section{Results}

\input{results.tex}

\section{Discussion and Future Work}

\subsection{Drawbacks and Future Plans}

Although our RBM-based anomaly detection system has proven an excellent
performance in the experiments, RBMs are just shadow generative networks
with two layers. The direct extension to deep generative networks
such DBNs \citep{Hinton.Geoffrey_etal_2006NeuralCompututation_DBN}
or DBMs \citep{Salakhutdinov.Ruslan_Hinton.Geoffrey_2009AISTATS}
offers more powerful capacity of encoding the normality distribution
and produces superior detection performance. In the future, we intend
to integrate the deep architecture into our anomaly framework and
discover effective mechanism to train and do inference in such multi-layer
generative nets. For further extension, we aim to develop a deep abnormality
detection system that is a deep generative network specialising in
the problem of anomaly detection, instead of adapting the popular
deep networks in literature that are designed for general purpose.

\subsection{Significance and Benefits}

Our research introduces an effective framework to detect anomaly events
in streaming surveillance videos. Broadly speaking, the significance
of our work lies on developing a powerful and generalized tool with
capacity of unsupervised learning, automatic representation and less
need for human-intervention that enables to isolate unexpected signals
and unusual patterns in data collection. This research also develops
a comprehensive understanding of the usage of generative networks
for anomaly detection. By analysing their strengths and drawbacks,
we can design more efficient and effective networks that specialize
to localize abnormal data points. However, the multi-layer architecture
in deep networks inspires us with the idea of hierarchical anomaly
detection systems where anomaly data (e.g. unusual pixels or regions
in video frames) are detected in the bottom layers and abstract anomaly
information (e.g. anomaly scenes in videos) is represented in the
top layers. Finally, many practical applications such as video analysis,
fraud detection, structural defect detection, medical anomaly detection
will benefit from our proposed system.

\section{Conclusion}

Throughout this report, we have described the current research trend
in the problem of anomaly detection, and more specifically, video
anomaly detection. We also have summarized the recent studies of generative
networks that are grounded in deep learning and neural network principles.
The literature review pointed out three key limitations of existing
anomaly detectors that are insufficient labeled data, the ambiguous
definition of abnormality and the costly step of feature representation.
To overcome these limitations, we have introduced our idea of utilising
the power of generative networks to model the regular data distribution
using restricted Boltzmann machines, and then detect anomaly events
via reconstruction errors. We conducted the experiments on three benchmark
datasets of UCSD Ped1, Ped2 and Avenue for video anomaly detection
and compared our network with several baselines. The experimental
results show that our shallow network can obtain the comparable performance
with the state-of-the-art anomaly detectors. In future work, we aim
to build a deep anomaly detection system that is specially designed
for anomaly detection. Many application domains, including video analysis
and scene understanding, can benefit from the results of our research. 

\bibliographystyle{plainnat}
\bibliography{hungv}

\end{document}

%% file: macros_hungv.tex
\selectlanguage{english}%

\global\long\def\bA{\boldsymbol{A}}

\global\long\def\bB{\boldsymbol{B}}

\global\long\def\bC{\boldsymbol{C}}

\global\long\def\bc{\boldsymbol{c}}

\global\long\def\norm{\left\Vert \right\Vert }

\global\long\def\realmn{\realset^{m\times n}}

\global\long\def\realm{\realset^{m}}

\global\long\def\bb{\boldsymbol{b}}

\global\long\def\rank{\text{rank}}

\global\long\def\lrr#1{\left(#1\right)}

\global\long\def\lrc#1{\left\{  #1\right\}  }

\global\long\def\lrs#1{\left[#1\right]}

\global\long\def\lrv#1{\left|#1\right|}

\global\long\def\mattwtw#1#2#3#4{\left[\begin{array}{rr}
 #1  &  #2\\
 #3  &  #4 
\end{array}\right]}

\global\long\def\mattrtr#1#2#3#4#5#6#7#8#9{\left[\begin{array}{rrr}
 #1  &  #2  &  #3\\
 #4  &  #5  &  #6\\
 #7  &  #8  &  #9 
\end{array}\right]}

\newcommand{\sidenote}[1]{\marginpar{\small \emph{\color{Medium}#1}}}

\global\long\def\se{\hat{\text{se}}}

\global\long\def\interior{\text{int}}

\global\long\def\boundary{\text{bd}}

\global\long\def\ML{\textsf{ML}}

\global\long\def\GML{\mathsf{GML}}

\global\long\def\HMM{\mathsf{HMM}}

\global\long\def\support{\text{supp}}

\global\long\def\new{\text{*}}

\global\long\def\stir{\text{Stirl}}

\global\long\def\mA{\mathcal{A}}

\global\long\def\mB{\mathcal{B}}

\global\long\def\mF{\mathcal{F}}

\global\long\def\mK{\mathcal{K}}

\global\long\def\mH{\mathcal{H}}

\global\long\def\mX{\mathcal{X}}

\global\long\def\mZ{\mathcal{Z}}

\global\long\def\mS{\mathcal{S}}

\global\long\def\Ical{\mathcal{I}}

\global\long\def\mT{\mathcal{T}}

\global\long\def\Pcal{\mathcal{P}}

\global\long\def\dist{d}

\global\long\def\HX{\entro\left(X\right)}
 \global\long\def\entropyX{\HX}

\global\long\def\HY{\entro\left(Y\right)}
 \global\long\def\entropyY{\HY}

\global\long\def\HXY{\entro\left(X,Y\right)}
 \global\long\def\entropyXY{\HXY}

\global\long\def\mutualXY{\mutual\left(X;Y\right)}
 \global\long\def\mutinfoXY{\mutualXY}

\global\long\def\given{\mid}

\global\long\def\gv{\given}

\global\long\def\goto{\rightarrow}

\global\long\def\asgoto{\stackrel{a.s.}{\longrightarrow}}

\global\long\def\pgoto{\stackrel{p}{\longrightarrow}}

\global\long\def\dgoto{\stackrel{d}{\longrightarrow}}

\global\long\def\lik{\mathcal{L}}

\global\long\def\logll{\mathit{l}}

\global\long\def\vectorize#1{\mathbf{#1}}

\global\long\def\vt#1{\mathbf{#1}}

\global\long\def\gvt#1{\boldsymbol{#1}}

\global\long\def\idp{\ \bot\negthickspace\negthickspace\bot\ }
 \global\long\def\cdp{\idp}

\global\long\def\das{:=}

\global\long\def\id{\mathbb{I}}

\global\long\def\idarg#1#2{\id\left\{  #1,#2\right\}  }

\global\long\def\iid{\stackrel{\text{iid}}{\sim}}

\global\long\def\bzero{\vt 0}

\global\long\def\bone{\mathbf{1}}

\global\long\def\boldm{\boldsymbol{m}}

\global\long\def\bff{\vt f}

\global\long\def\bx{\boldsymbol{x}}

\global\long\def\bl{\boldsymbol{l}}

\global\long\def\bu{\boldsymbol{u}}

\global\long\def\ba{\boldsymbol{a}}

\global\long\def\bo{\boldsymbol{o}}

\global\long\def\bh{\boldsymbol{h}}

\global\long\def\bs{\boldsymbol{s}}

\global\long\def\bz{\boldsymbol{z}}

\global\long\def\xnew{y}

\global\long\def\bxnew{\boldsymbol{y}}

\global\long\def\bX{\boldsymbol{X}}

\global\long\def\tbx{\tilde{\bx}}

\global\long\def\by{\boldsymbol{y}}

\global\long\def\bY{\boldsymbol{Y}}

\global\long\def\bZ{\boldsymbol{Z}}

\global\long\def\bU{\boldsymbol{U}}

\global\long\def\bv{\boldsymbol{v}}

\global\long\def\bn{\boldsymbol{n}}

\global\long\def\bV{\boldsymbol{V}}

\global\long\def\bI{\boldsymbol{I}}

\global\long\def\bw{\vt w}

\global\long\def\balpha{\gvt{\alpha}}

\global\long\def\bbeta{\gvt{\beta}}

\global\long\def\bmu{\gvt{\mu}}

\global\long\def\btheta{\boldsymbol{\theta}}

\global\long\def\blambda{\boldsymbol{\lambda}}

\global\long\def\bgamma{\boldsymbol{\gamma}}

\global\long\def\bpsi{\boldsymbol{\psi}}

\global\long\def\bphi{\boldsymbol{\phi}}

\global\long\def\bpi{\boldsymbol{\pi}}

\global\long\def\bomega{\boldsymbol{\omega}}

\global\long\def\bepsilon{\boldsymbol{\epsilon}}

\global\long\def\btau{\boldsymbol{\tau}}

\global\long\def\realset{\mathbb{R}}

\global\long\def\realn{\realset^{n}}

\global\long\def\integerset{\mathbb{Z}}

\global\long\def\natset{\integerset}

\global\long\def\integer{\integerset}

\global\long\def\natn{\natset^{n}}

\global\long\def\rational{\mathbb{Q}}

\global\long\def\rationaln{\rational^{n}}

\global\long\def\complexset{\mathbb{C}}

\global\long\def\comp{\complexset}

\global\long\def\compl#1{#1^{\text{c}}}

\global\long\def\and{\cap}

\global\long\def\compn{\comp^{n}}

\global\long\def\comb#1#2{\left({#1\atop #2}\right) }

\global\long\def\nchoosek#1#2{\left({#1\atop #2}\right)}

\global\long\def\param{\vt w}

\global\long\def\Param{\Theta}

\global\long\def\meanparam{\gvt{\mu}}

\global\long\def\Meanparam{\mathcal{M}}

\global\long\def\meanmap{\mathbf{m}}

\global\long\def\logpart{A}

\global\long\def\simplex{\Delta}

\global\long\def\simplexn{\simplex^{n}}

\global\long\def\dirproc{\text{DP}}

\global\long\def\ggproc{\text{GG}}

\global\long\def\DP{\text{DP}}

\global\long\def\ndp{\text{nDP}}

\global\long\def\hdp{\text{HDP}}

\global\long\def\gempdf{\text{GEM}}

\global\long\def\rfs{\text{RFS}}

\global\long\def\bernrfs{\text{BernoulliRFS}}

\global\long\def\poissrfs{\text{PoissonRFS}}

\global\long\def\grad{\gradient}
 \global\long\def\gradient{\nabla}

\global\long\def\partdev#1#2{\partialdev{#1}{#2}}
 \global\long\def\partialdev#1#2{\frac{\partial#1}{\partial#2}}

\global\long\def\partddev#1#2{\partialdevdev{#1}{#2}}
 \global\long\def\partialdevdev#1#2{\frac{\partial^{2}#1}{\partial#2\partial#2^{\top}}}

\global\long\def\closure{\text{cl}}

\global\long\def\cpr#1#2{\Pr\left(#1\ |\ #2\right)}

\global\long\def\var{\text{Var}}

\global\long\def\Var#1{\text{Var}\left[#1\right]}

\global\long\def\cov{\text{Cov}}

\global\long\def\Cov#1{\cov\left[ #1 \right]}

\global\long\def\COV#1#2{\underset{#2}{\cov}\left[ #1 \right]}

\global\long\def\corr{\text{Corr}}

\global\long\def\sst{\text{T}}

\global\long\def\SST{\sst}

\global\long\def\ess{\mathbb{E}}

\global\long\def\Ess#1{\ess\left[#1\right]}

\newcommandx\ESS[2][usedefault, addprefix=\global, 1=]{\underset{#2}{\ess}\left[#1\right]}

\global\long\def\fisher{\mathcal{F}}

\global\long\def\bfield{\mathcal{B}}
 \global\long\def\borel{\mathcal{B}}

\global\long\def\bernpdf{\text{Bernoulli}}

\global\long\def\betapdf{\text{Beta}}

\global\long\def\dirpdf{\text{Dir}}

\global\long\def\gammapdf{\text{Gamma}}

\global\long\def\gaussden#1#2{\text{Normal}\left(#1, #2 \right) }

\global\long\def\gauss{\mathbf{N}}

\global\long\def\gausspdf#1#2#3{\text{Normal}\left( #1 \lcabra{#2, #3}\right) }

\global\long\def\multpdf{\text{Mult}}

\global\long\def\poiss{\text{Pois}}

\global\long\def\poissonpdf{\text{Poisson}}

\global\long\def\pgpdf{\text{PG}}

\global\long\def\wshpdf{\text{Wish}}

\global\long\def\iwshpdf{\text{InvWish}}

\global\long\def\nwpdf{\text{NW}}

\global\long\def\niwpdf{\text{NIW}}

\global\long\def\studentpdf{\text{Student}}

\global\long\def\unipdf{\text{Uni}}

\global\long\def\transp#1{\transpose{#1}}
 \global\long\def\transpose#1{#1^{\mathsf{T}}}

\global\long\def\mgt{\succ}

\global\long\def\mge{\succeq}

\global\long\def\idenmat{\mathbf{I}}

\global\long\def\trace{\mathrm{tr}}

\global\long\def\argmax#1{\underset{_{#1}}{\text{argmax}} }

\global\long\def\argmin#1{\underset{_{#1}}{\text{argmin}\ } }

\global\long\def\diag{\text{diag}}

\global\long\def\norm{}

\global\long\def\spn{\text{span}}

\global\long\def\vtspace{\mathcal{V}}

\global\long\def\field{\mathcal{F}}
 \global\long\def\ffield{\mathcal{F}}

\global\long\def\inner#1#2{\left\langle #1,#2\right\rangle }
 \global\long\def\iprod#1#2{\inner{#1}{#2}}

\global\long\def\dprod#1#2{#1 \cdot#2}

\global\long\def\norm#1{\left\Vert #1\right\Vert }

\global\long\def\entro{\mathbb{H}}

\global\long\def\entropy{\mathbb{H}}

\global\long\def\Entro#1{\entro\left[#1\right]}

\global\long\def\Entropy#1{\Entro{#1}}

\global\long\def\mutinfo{\mathbb{I}}

\global\long\def\relH{\mathit{D}}

\global\long\def\reldiv#1#2{\relH\left(#1||#2\right)}

\global\long\def\KL{KL}

\global\long\def\KLdiv#1#2{\KL\left(#1\parallel#2\right)}
 \global\long\def\KLdivergence#1#2{\KL\left(#1\ \parallel\ #2\right)}

\global\long\def\crossH{\mathcal{C}}
 \global\long\def\crossentropy{\mathcal{C}}

\global\long\def\crossHxy#1#2{\crossentropy\left(#1\parallel#2\right)}

\global\long\def\breg{\text{BD}}

\global\long\def\lcabra#1{\left|#1\right.}

\global\long\def\lbra#1{\lcabra{#1}}

\global\long\def\rcabra#1{\left.#1\right|}

\global\long\def\rbra#1{\rcabra{#1}}

\global\long\def\likelihood{\mathcal{L}}

\global\long\def\normal{\mathcal{N}}

\global\long\def\bSigma{\boldsymbol{\Sigma}}

\global\long\def\dataset{\mathcal{D}}
\selectlanguage{american}%

%% file: coverpage.tex
\selectlanguage{english}%

\title{\includegraphics[width=1\columnwidth]{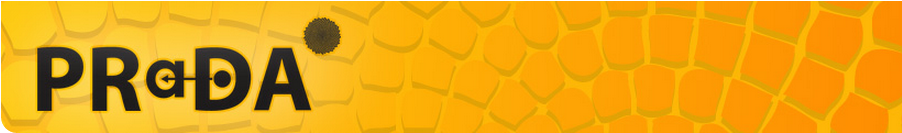}\vfill{}
\textbf{Energy-based Models for Video Anomaly Detection}}

\maketitle
\vspace{20pt}

\begin{doublespace}
\begin{center}
\textbf{\large{}Hung Vu}{\large{}$\dagger$}\textbf{\large{}, Dinh
Phung}{\large{}$\dagger$}\textbf{\large{}, Tu Dinh Nguyen}{\large{}$\dagger$}\textbf{\large{},
Anthony Trevors}{\large{}$\ddagger$}\textbf{\large{} and}\\
\textbf{\large{}Svetha Venkatesh}{\large{}$\dagger$}\textbf{\large{}
}{\large{}}\\
\vspace{10pt}
{\large{}$\dagger$Centre for Pattern Recognition and Data Analytics
(PRaDA)}\\
{\large{}School of Information Technology, Deakin University, Geelong,
Australia}\\
{\large{}Email: \{hungv, tu.nguyen, svetha.venkatesh, dinh.phung\}@deakin.edu.au
}\\
{\large{}$\ddagger$Defence Science and Technology Organization (DSTO),
Melbourne, Australia}\\
{\large{}Email: Anthony.Travers@dsto.defence.gov.au}
\par\end{center}{\large \par}
\end{doublespace}

\vfill{}

\begin{tabular}{>{\raggedright}b{0.65\paperwidth}c}
\textsc{\textcolor{blue}{\LARGE{}P}}\textsc{\LARGE{}attern }\textsc{\textcolor{blue}{\LARGE{}R}}\textsc{\LARGE{}ecognition
}\textsc{\textcolor{blue}{\noun{\LARGE{}a}}}\textsc{\LARGE{}nd }\textsc{\textcolor{blue}{\LARGE{}D}}\textsc{\LARGE{}ata
}\textsc{\textcolor{blue}{\LARGE{}A}}\textsc{\LARGE{}nalytics}\\
School of Information Technology, Deakin University, Australia.\\
Locked Bag 20000, Geelong VIC 3220, Australia.\\
Tel: +61 3 5227 2150 \\
Internal report number: \textsf{\textcolor{blue}{\small{}TR-PRaDA-01/17}},
April, 2017. & \includegraphics[scale=0.15]{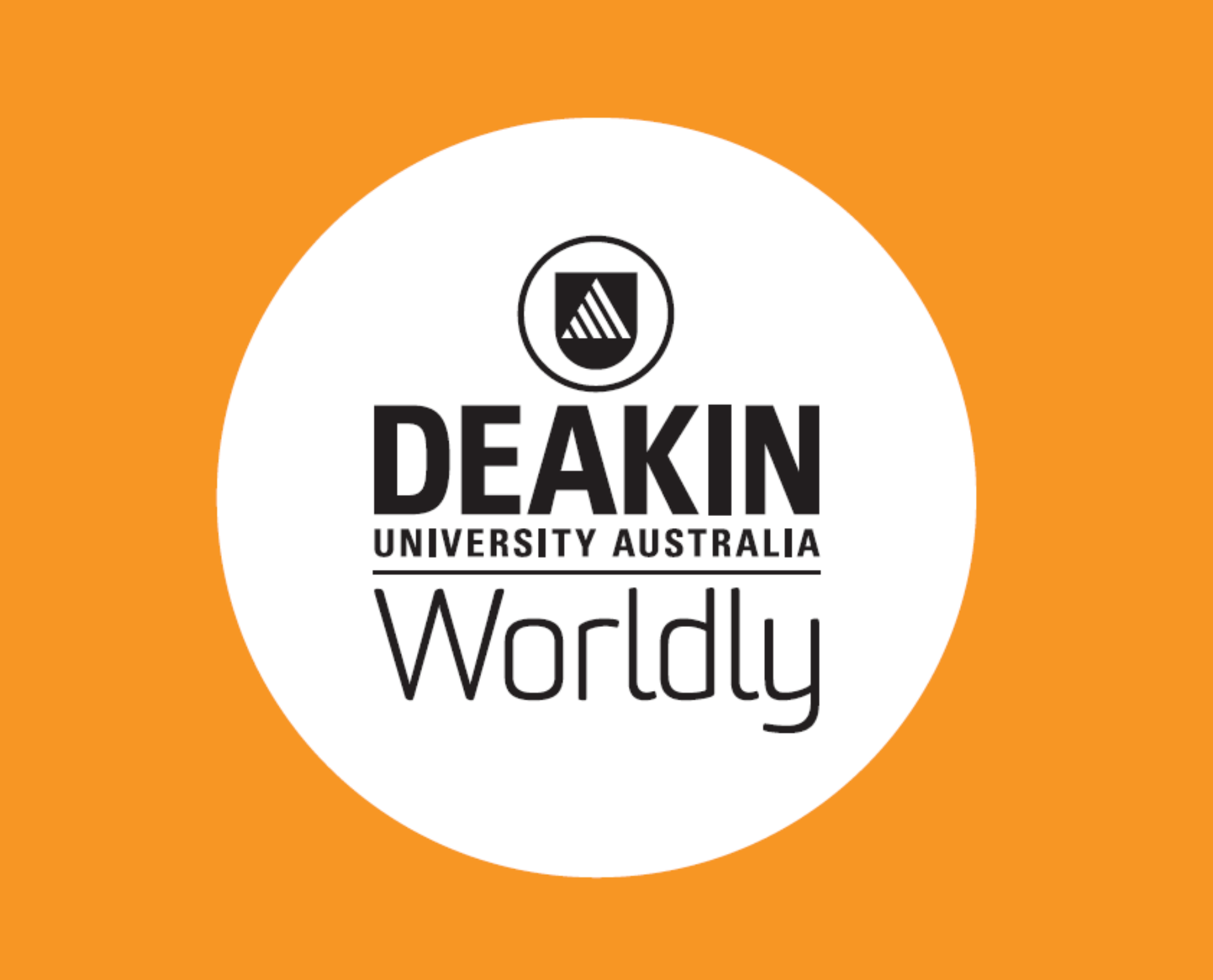}\tabularnewline
\end{tabular}

\thispagestyle{empty}\selectlanguage{american}%

%% file: intro.tex
Anomaly detection is one of the most important problems that has been
attracting intensive research interest in recent years \citep{Pimentel.Marco_etal_2014SignProcess_Novelty_Review}.
 Anomaly detection systems cover a wide range of domains and topics
such as wrong behaviour detection and traffic management in computer
vision, intrusion detection in computer systems and networking, credit
card and insurance claim fraud detection in daily activities, disease
detection in healthcare. Although the definition of anomaly detection
varies across areas and systems, formally, it usually refers to ``the
problem of finding patterns in data that do not conform to expected
behaviours'' \citep{Chandola.Varun_etal_2009_review}. The expected
patterns or behaviours are mentioned as ``normal'' whilst the non-conforming
ones are known as ``abnormal'' or ``anomaly''. Building an effective
anomaly system, which completely distinguishes abnormal data points
from normal regions, is a non-trivial task. Firstly, due to the complexity
of real data, anomaly data points may lie closely to the boundary
of normal regions, e.g. skateboarders and walking people appear similarly
in the application of camera surveillance, where skateboarders are
anomaly objects and prohibited in pedestrian footpaths. Secondly,
abnormal objects can evolve over time and possibly become normal,
e.g. high temperature is undesired in winters but is normal in summers.
Another difficulty is the shortage of labelled data. More specifically,
the normal patterns are usually available or easy to be collected,
e.g. healthy patient records, but the abnormal samples are relatively
few or costly, e.g. the problem of jet engine failure detection requires
to destroy as many engines as possible for abnormal data acquisition.
According to \citep{Pimentel.Marco_etal_2014SignProcess_Novelty_Review},
anomaly detection methods are classified into five main approaches.

\begin{itemize}
\item \textbf{Probabilistic approach}: This approach builds a model to estimate
the probability distribution that is assumed to generate the training
data. The samples with low probabilities, below a pre-defined threshold,
are considered as abnormality.
\item \textbf{Distance-based approach}: By assuming that the normal samples
occur in a dense neighbourhood while anomaly data is far from its
neighbours, nearest neighbour or clustering-based techniques can be
used to measure the similarity between data points and then isolate
abnormalities. 
\item \textbf{Domain-based approach}: Methods such as SVDD or One-class
SVM learn the boundaries of target classes from the training dataset.
The test data are identified as normal or abnormal based on their
locations with respect to the learned boundaries. 
\item \textbf{Information theoretic approach}: These methods accept the
idea that anomalies considerably change the information content of
the whole dataset. Therefore, an ensemble of points can be determined
as anomaly if its removal from the dataset makes a big change in information
content. 
\item \textbf{Reconstruction-based approach}: This approach struggles to
encode the data in a compact way but still maintain the significant
variability of the data. The abnormal data usually creates high reconstruction
errors that are the distance between the original data and the reconstructed
data. This approach consists of groups of neural network-based methods
and subspace-based methods. 
\end{itemize}
Most of current anomaly detection methods above rely on the feature
extraction steps to find a mapping from input space to feature space
where they believe that the new representation of data facilitates
the recognition of anomaly data points. Nevertheless, due to the complexity
of data in reality, a good feature representation not only requires
human-intensive efforts in design but also massive trails and failures
in experiments. By learning the data distribution, we avoid depending
on feature extraction steps and we can utilize the availability of
unlabeled normal data. Under this view, our method falls into the
first approach of probability. However, differing from other conventional
probabilistic generative methods such as GMM, we leverage on generative
neural networks, i.e. restricted Boltzmann machines, to power the
capacity of modeling complicated data distribution. The success of
deep learning techniques in solving many challenging problems such
as speech recognition \citep{Pascanu.Razvan_etal_2014ICLR_app_speech,Chung.Junyoung_etal_2014NIPS_app_speech},
object recognition \citep{Krizhevsky.Alex_etal_2012NIPS_app_objectrecognition},
pedestrian detection \citep{Sermanet.Pierre_etal_2013CVPR_app_pedestriandetection},
scene labeling \citep{Farabet.Clement_etal_2013PAMI_app_scenelabling}
encourages our research on generative networks for anomaly detection
in general and video anomaly detection problem in particular.

%% file: literature_review.tex
In this section, we firstly provide an overview of neural networks,
especially deep architecture, which have developed recently in literature.
Since our purpose is to model the probability distribution of normal
data, we focus on generative networks that are able to generate data
samples. The advantages and disadvantages of these networks are also
discussed in details. To demonstrate a practical application of abnormality
detection system, we review some work to solve the problem of abnormality
detection in video data.

\subsection{Generative Models}

In general, generative neural architectures are categorized into three
approaches, based on the type of connections between units, that are
directed networks, undirected networks and hybrid networks.

\subsubsection{Directed Generative Nets}

Sigmoid belief networks (SBNs) \citep{Neal.RadfordM_1990TR_SigmoidBeliefNet}
(also known as connectionist networks) are a class of Bayesian networks
that are directed acyclic graphs (DAGs). SBNs define activation functions
as logistic sigmoids. A node $X_{i}$ and its incoming edges encode
the local conditional probability of $X_{i}$ given its parent nodes,
denoted $\pi_{i}$. The joint probability over the network is defined
as the factorisation of these local conditional probabilities. 
\[
p(\bX=\bx)=\prod_{i=1}^{M}p\lrr{X_{i}=x_{i}|\bX_{\pi_{i}}=\bx_{\pi_{i}}}
\]

However, it is problematic to compute the conditional probability
distribution since exact inference over hidden variables in sigmoid
belief nets is intractable. Since samples can be drawn efficiently
from the network via ancestral sampling in the parent-child order,
inference by Gibbs sampling is possible in \citep{Neal.RadfordM_1992_SigmoidBeliefNetwork}
but just in small networks of tens of units. Consequently, inference
becomes the central problem in the sigmoid belief networks. \citet{Saul.LawrenceK_etal_1996_SigmoidBeliefNet_variational}
and \citet{Saul.Lawrence_Jordan.Michael_1999_SigmoidBeliefNetwork}
struggle to use variational distributions to estimate the inference.
The parameters of the variational mean-field distribution can be learned
automatically using the wake-sleep algorithm in Helmholtz machines
\citep{Dayan.Peter_etal_1995NeuralComputation_SigmoidBeliefNet_learnedinference,Dayan.Peter_Hinton.Geoffrey_1996_SigmoidBeliefNet_learnedinference}.
Recently, several proposed methods such as reweighted wake-sleep \citep{BornscheinJorg_BengioYoshua_2015ICLR_SigmoidBeliefNet_reweightedwakesleep}
and bidirectional Helmholtz machines \citep{Bornsche.Jorg_etal_2016ICML_SigmoidBeliefNet_biHelmholtzMachines}
can help to speed up the training procedure as well as obtain the
state-of-the-art performance \citep{BornscheinJorg_BengioYoshua_2015ICLR_SigmoidBeliefNet_reweightedwakesleep,Bornsche.Jorg_etal_2016ICML_SigmoidBeliefNet_biHelmholtzMachines}.

The directed models mentioned above have two main drawbacks. Firstly,
they usually work well only on discrete latent variables. Most networks
\citep{Pascanu.Razvan_etal_2014ICLR_app_speech,Krizhevsky.Alex_etal_2012NIPS_app_objectrecognition}
prefer real-valued latent units, which can describle more accurately
the complex data, e.g. photos or voices, in practical applications.
Secondly, these sigmoid belief networks with variational inference
assume a fully factorial mean-field form for their approximate distribution
but such assumption is too strict. Some studies \citep{Kingma.Diederik_2013arxiv_VAE,Rezende.DaniloJimenez_2014ICML_VAE,Kingma.Diederik_Welling.Max_2014ICLR_VAE}
relax this constraint by reparameterising the variational lower bound
into a differentiable function which can be trained efficiently with
gradient ascent training methods. Surprisingly, this trick restates
the problem of training the directed graphical model as training an
autoencoder. The network with the capability of optimizing the variational
parameters via a neural network is known as a variational autoencoder
\citep{Kingma.Diederik_2013arxiv_VAE,Rezende.DaniloJimenez_2014ICML_VAE,Kingma.Diederik_Welling.Max_2014ICLR_VAE}.
Data generation in this network is done similarly to other directed
graphical models via ancestral sampling. However, since variational
autoencoders require differentiation over the hidden variables, they
cannot originally work with discrete latent variables. The idea of
variational autoencoders was extended to sequential data in variational
RNN \citep{Chung.Junyoung_eltal_2015NIPS_VariationalRNN} and discrete
data in discrete variational autoencoders \citep{Rolfe.JasonTyler_2016arxiv}.
Another variant of variational autoencoder, whose log likelihood lower
bound is derived from importance weighting, was introduced in \citep{Burda.Yuri_etal_2015_VAE_ImportanceWeightedAE}. 

Another direction to avoid the inference problem in directed graphical
models is generative adversarial networks (GANs) \citep{Goodfellow.Ian_etal_2014NIPS_generativeadversarialnetwork}.
GANs sidestep the difficulty by viewing the network training as the
game theoretic problem where two neural networks compete with each
other in a minimax two-player game. To be more precise, a GAN simultaneously
trains two neural networks: a differential generator network $G\lrr{\bz;\btheta_{D}}$,
which aims to capture the data distribution, and its adversary, a
discriminative network $D\lrr{\bx;\btheta_{D}}$, which estimates
the probability of a sample coming from training data. Conceptually,
$G$ learns to fool $D$ while $D$ attempts to distinguish the samples
from the training data distribution and the fake samples from the
generator $G$. To be more precise, $G$ adapts itself to make $D\lrr{G\lrr{\bz,\btheta_{G}},\btheta_{D}}$
close to 1 as much as possible while $D$ attemps to produce high
probability $D\lrr{\bx}\approx1$ for data samples and low probability
$D\lrr{\tilde{\bx},\btheta_{D}}\approx0$ for generated data $\tilde{\bx}=G\lrr{\bz,\btheta_{G}}$.
At convergence, we expect that the generator can totally deceive the
discriminator while the discriminator produces $\nicefrac{1}{2}$
everywhere. This results in a minimax game with the value function
described as: 
\[
\min_{\btheta_{G}}\max_{\btheta_{D}}f\lrr{\bx;G,D}
\]
where $f\lrr{\bx;\btheta_{G},\btheta_{D}}=\ess_{p_{\text{data }}\lrr{\bx}}\lrs{\log D\lrr{\bx;\btheta_{D}}}+\ess_{p_{\bz}\lrr{\bz}}\lrs{\log\lrr{1-D\lrr{G\lrr{\bz;\btheta_{G}}}}}$.

The training procedure is run efficiently with gradient ascent method.
In the case where $f\lrr{\bx;\btheta_{G},\btheta_{D}}$ is convex,
the generative distribution can converge to the data distribution.
In general, it is not guaranteed that the training reaches at equilibrium.
GAN offers many advantages including no need for Markov chains and
inference estimation, training with back-propagation, a variety of
functions integrated into the network. Its drawback is that it is
unable to describe data probability distribution $p_{G}\lrr{\bx;\btheta_{G}}$
explicitly. Some extensions of GANs include the conditional versions
on class labels $p_{G}\lrr{\bx|\by;\btheta_{G}}$ \citep{Mirza.Mehdi_Osindero.Simon_2014_conditionalGAN},
a series of conditional GANs each of which generates images at a scale
of Laplacian pyramid \citep{Denton.EmilyL_etal_2015NIPS_LAPGAN} and
deep convolutional GANs \citep{Radford.Alec_2015_DeepConvGAN} for
image synthesis.

When working on image data, it is more useful to integrate convolutional
structure. Convolutional generative networks are deep networks based
on convolution and pooling operations, similar to convolutional neural
networks (CNNs) \citep{Lecun.Yann_etal_1998_CNN}, to generate images.
In the recognition network, the information flows from the bottommost
layer of images to intermediate convolution and pooling layers and
ends at the top of the network, often class labels. By contrast, the
generator network propagates signals in the reversed order to generate
images from the provided values of the top units. The mechanism for
image generation is to invert the pooling layers which are reduction
operations and are known as non-invertible. \citet{Dosovitskiy.Alexey_etal_2015CVPR_generativeCNN}
performs unpooling by replacing each value in a feature map by a block,
whose top-left corner is that value and other entries are zeros. Thought
this ``inverse'' operation is incorrect theoretically, the system
works fair well to generate relatively high-quality images of chairs
given an object type, viewpoint and colour. Recently, the convolutional
structure have also been combined with other deep learning networks
to form convolutional variational autoencoders \citep{Pu.Yunchen_etal_2016NIPS_convVAEs}
or deep convolutional generative adversarial networks (DCGANs) \citep{Radford.Alec_2015_DeepConvGAN}.

\subsubsection{Undirected Generative Nets}

Another graphical language to express probability distributions is
undirected graphical models, otherwise known as Markov random fieldd
or Markov networks \citep{Koller.Daphne_Friedman.Nir_2009_PGM}, which
use undirected edges to represent the interactions between random
variables. Undirected graphs are useful in the cases where causality
relationships are stated unclearly or there are no one-directional
interactions between identities. The primary methods in this direction
usually accept an energy-based representation where the joint probability
distribution is defined as
\[
p\lrr{\bv,\bh;\bpsi}=\frac{e^{-E\lrr{\bv,\bh;\bpsi}}}{\partfunc\lrr{\bpsi}}
\]
where $\bv$ and $\bh$ are observed and hidden variables respectively
and $\partfunc\lrr{\bpsi}=\sum_{\bv,\bh}e^{-E\lrr{\bv,\bh;\bpsi}}$,
named partition function, is the sum of unnormalized probability $\tilde{p}\lrr{\bv,\bh;\bpsi}=e^{-E\lrr{\bv,\bh;\bpsi}}$
of all possible discrete states of the networks. Like most probabilistic
frameworks, our purpose is to maximize the log likelihood function,
that is: 
\[
\log p\lrr{\bv;\bpsi}=\log\sum_{\bh}e^{-E\lrr{\bv,\bh;\bpsi}}-\log\partfunc\lrr{\bpsi}
\]

The favourite learning scheme for maximum likelihood is gradient ascent
where its log likelihood gradient is reformulated as the sum of expectations:
\begin{eqnarray}
\grad_{\bpsi}\log p\lrr{\bv;\bpsi} & = & \ess_{p\lrr{\bh|\bv;\bpsi}}\lrs{\grad_{\bpsi}\log\tilde{p}\lrr{\bv,\bh;\bpsi}}-\ess_{p\lrr{\bv,\bh;\bpsi}}\lrs{\grad_{\bpsi}\log\tilde{p}\lrr{\bv,\bh;\bpsi}}\label{eq:UGM_gradient}
\end{eqnarray}

The first expectation is called the positive phase that depends on
the posterior distribution given the training data samples while the
latter is known as the negative phase that are related to samples
drawn from the model distribution. Both are computationally intractable
in general. To be more precise, the exact computation of positive
phase takes time that is exponential in the number of hidden units
while the cost of the second phase exponentially increases with respect
to the number of units in the model. Undirected generative methods
differ in treating these terms. We start with introducing Boltzmann
machines that confront both difficulties simultaneously.

A Boltzmann machine \citep{Fahlman.Scott_etal_1983_BM,Ackley.DavidH_etal_1985_BM,Hinton.Geoffrey_etal_1984_BM,Hinton.Geoffrey_Sejnowski.Terrence.Joseph_1986_BM}
consists of a group of visible and hidden units that freely interact
with each other. Its energy function is defined as
\begin{eqnarray}
E\lrr{\bv,\bh,\bpsi} & = & -\frac{1}{2}\bv^{\top}\bO\bv-\frac{1}{2}\bh^{\top}\bU\bh-\bv^{\top}\bW\bh-\ba^{\top}\bv-\bb^{\top}\bh\label{eq:BM_energy_function}
\end{eqnarray}
wherein $\bO,\bU$ and $\bW$ are weight matrices that encode the
strength of visible-to-visible, hidden-to-hidden and visible-to-hidden
interactions respectively. Basically, Boltzmann machines follow the
log likelihood gradient expression in Eq. \ref{eq:UGM_gradient}.
This means that the explicit computation of the gradient is infeasible,
and hence the Boltzmann machine learning depends on approximation
techniques. Hinton and Sejnowski \citep{Hinton.Geoffrey_Sejnowski.TerrenceJoseph_1983CVPR_BM_Gibbssampling}
used two Gibbs chains \citep{Geman.Stuart_Geman.Donald_1984_GibbsSampling}
to estimate two expectations independently. The drawback of this method
is the low mixing rate to reach the stationary distribution since
Gibbs sampling needs much time to discover the landscape of the highly
multimodal energy function. Some methods such as stochastic maximum
likelihood \citep{Neal.RadfordM_1992_SigmoidBeliefNetwork,Younes.Laurent_1989_BM_converence,Yuille.Alan_2004_BM_convergence}
improve the speed of convergence by reusing the final state of the
previous Markov chain to initialize the next chain. Boltzmann machines
can also be learned with variational approximation \citep{Hinton.Geoffrey_Zemel.RichardStanley_1993NIPS_variational,Neal.RadfordM_Hinton.Geoffrey_1999_variational,Jordan.Michael_etal_1999_variational}.
The true complex posterior distribution $p\lrr{\bh|\bv;\bpsi}$ of
Boltzmann machines is replaced by a simpler approximate distribution
$q\lrr{\bh|\bv;\btheta}$ whose parameters are estimated via maximizing
the lower bound on the log likelihood. The popular procedure to train
Boltzmann machines usually combines both stochastic maximum likelihood
and variational mean-field assumption $q\lrr{\bh|\bv;\btheta}=\prod_{i}q\lrr{h_{i}|\bv;\btheta}$
\citep{Salakhutdinov.Ruslan_2009thesis}.

Although there are several proposed methods to train general Boltzmann
machines, they are unable to work effectively in huge networks. By
simplifying the graph structure and putting more constraints on network
connections, we can end up with an easy-to-train Boltzmann machines.
One quintessence of such networks is restricted Boltzmann machines
(RBMs), also known under the name harmonium \citep{Smolensky_1986},
that remove all visible-to-visible and hidden-to-hidden connections.
In other words, RBMs have two layers, one of hidden variables and
the other of visible variables, and there are no connections between
units in the same layers. Interestingly, this ``restriction'' comes
up with a nice property, which is all units in the same layer are
conditionally independent given the other layer. Consequently, the
positive phase can be analytically computed in RBMs while the negative
phase can be estimated efficiently with contrastive divergence \citep{Hinton.Geoffrey_2002NeuralComp_CD}
or persistent contrastive divergence \citep{Tieleman.Tijmen_2008ICML_PCD},
also termed as stochastic maximum likelihood, by sampling from the
model distribution via alternative Gibbs samplers. Despite the smaller
number of interactions between variables, binary RBMs are justified
to be powerful enough to approximate any discrete distribution \citep{LeRoux.Nicolas_Bengio.Yoshua_2008NeuralComp_theory_addhidden}.
They are also effective tools for unsupervised learning and representation
learning \citep{Nguyen.TuDinh_etal_2013PAKDD_Mv.RBM_latenpatient,Nguyen.TuDinh_eltal_2013ACML_NRBM,Tran.Truyen_etal_2015JournalBI_eNRBM}. 

RBMs have been extended in different directions in the last decades.
In 2005, Welling et al. \citep{Welling.Max_etal_2005NIPS_exponential}
introduced a generalized RBM for many exponential family distributions.
Most studies on RBMs develop the different versions of RBMs for a
variety of data types. For examples, methods such as mean and covariance
RBMs (mcRBMs) \citep{Ranzato.MarcAurelio_etal_2010AISTATS_mcRBM},
the mean-product of student's t-distribution models (mPoT models)
\citep{Ranzato.MarcAurelio_etal_2010NIPS_mPoT} and spike and slab
RBMs (ssRBMs) \citep{Courville.Aarron_etal_2011ICML_ssRBM} focus
on studying the capacity of RBMs to deal with real-valued data. For
image data, convolutional structure is also integrated into RBMs in
\citep{Desjardins.Guillaume_Bengio.Yoshua_2008TR_convRBM}. Sequential
data can be modeled by conditional RBMs in \citep{Taylor.GrahamW_etal_2006NIPS_RBM_app}
which learn $p\lrr{\bx^{t}|\bx^{t-1},\ldots,\bx^{t-m}}$ from the
sequence of joint angles of human skeletons and then are able to generate
3D motions. Other variants of conditional RBMs are HashCRBMs \citep{Mnih.Volodymyr_etal_2011UAI_conditionalRBM}
and RNNRBM \citep{BoulangerLewandowski.Nicolas_etal_2012ICML_conditionalRBM_app}.
It is possible to modify RBMs to model $p\lrr{\by|\bx}$, for example
\citep{Larochelle.Hugo_Bengio.Yoshua_2008ICML_discriminativeRBM},
where RBMs can work in both generative and discriminative manners.
However, the discriminative RBMs do not seem to be superior traditional
classifiers such as multi-layer perceptrons (MLPs). Due to its effectiveness
and power, RBMs are integrated into many systems including collaborative
filtering \citep{Salakhutdinov.Ruslan_etal_2017ICML_RBM_app}, information
and image retrieval \citep{Gehler_etal_2006ICML_RBM_app} and time
series modeling \citep{Sutskever.Ilya_Hinton.Geoffrey_2007AISTATS_conditionalRBM_app,Taylor.GrahamW_etal_2006NIPS_RBM_app}.

Since RBMs with single hidden layer have limitations to represent
features hierarchically, one is interested in deep networks whose
deeper layers describe higher-level concepts. Furthermore, deep architectures
need smaller models than shallow networks to represent functions \citep{Larochelle.Hugo_Bengio.Yoshua_2008ICML_discriminativeRBM}.
The circuit complexity theory shows that the deep circuits are more
exponentially efficient than the shallow ones \citep{Ajtai.M_1983_deeplearning_better_deepcircuit,Hastad.Johan_1987Book_deeplearning_better_deepcircuit,Allender.Eric_1996_deeplearning_better_deepcircuit}.
This results in the invention of deep Boltzmann machines. Essentially,
deep Boltzmann machines (DBMs) \citep{Salakhutdinov.Ruslan_Hinton.Geoffrey_2009AISTATS}
are Boltzmann machines whose units are organized in multiple layer
networks without intra-layer connections. Like BMs, the log likelihood
function is a computational challenge because of the presence of two
intractable expectations. Sampling in DBMs is not efficient and it
requires the involvement of most units in the graph. Meanwhile, it
is difficult to tackle the intractable posterior distribution unless
it is approximated by variational inference. Moreover, there is the
need for a layer-wise pretraining stage \citep{Salakhutdinov.Ruslan_Hinton.Geoffrey_2009AISTATS}
to move the parameters to good values in parameter space in order
to successfully train DBM models. After pretraining initialization,
DBMs can be jointly trained with general BM training procedure, e.g.
one proposed in \citep{Salakhutdinov.Ruslan_2009thesis}. Alternatively,
DBM models can be learned, without pretraining, using some methods
such as centred DBMs \citep{Montavon.Gregoire_Muller.KlauseRobert_2012NN_CenterDBM}
and multi-prediction DBMs (MPDBMs) \citep{Goodfellow.Ian_etal_2013NIPS_MPDBM}.
In centred DBMs, the energy function is rewritten as the function
of centred states $\lrr{\bv-\balpha,\bh-\bbeta},$ where $\balpha$
and $\bbeta$ are offsets associated with visible or hidden units
of the network, instead of states $\lrr{\bv,\bh}$ as usual. This
trick leads to a better conditioned optimization formula, in terms
of a smaller ratio of the largest and lowest eigenvalues of the Hessian
matrix, and then it enables the learning procedure to be more stable
and robust to noise. The centring trick is known to be equivalent
to enhanced gradients, introduced by \citet{Cho.KyungHyun_etal_2011ICML_enhancedgradient}.
The second method of MPDBMs \citep{Goodfellow.Ian_etal_2013NIPS_MPDBM}
offers an alternative criterion to train DBM without maximizing the
likelihood function. This criterion is the sum of terms that describe
the ability of the model to infer the values of a subset of observed
variables given the other observed variables. Interestingly, this
can be viewed as training a family of recurrent neural networks, which
share the same parameters but solve various inference tasks. As a
result, MPDBMs allow to be trained with back-propagation and then
avoid confronting MCMC estimate of the gradient. Since MPDPMs are
designed to focus on handling inference tasks, the ability of generating
realistic samples is not good. Overall, MPDPMs can be viewed to train
models to maximize a variational approximation to generalized pseudolikelihood
\citep{Besag.Julian_1975_BM_pseudolikelihood}. 

All aforementioned BMs encode two-way interactions between two variables
in the network, for example unit-to-unit interaction terms in Eq.
\ref{eq:BM_energy_function}. One research direction is to discover
higher-order Boltzmann machines \citep{Sejnowski.TerrenceJ_1986NN_HigherOrderBM}
whose energy functions include the product of many variables. \citet{Memisevic.Roland_Hinton.Geoffrey_2007CVPR_HigherOrderBM,Memisevic.Roland_Hinton.Geoffrey_2010_HigherOrderBM}
proposed Boltzmann machines with third-order connections between a
hidden unit and a pair of images to model the linear transformation
between two input images or two consecutive frames in videos. A hidden
unit and a visible unit can communicate with a class label variable
to train discriminative RBMs for classification \citep{Luo.Heng_etal_2011AISTATS_HigherOrderRBM}.
\citet{Sohn.Kihyuk_etal_2013ICML_HigherOrderBM} introduced a higher-order
Boltzmann machine with three-way interactions with masking variables
that turn on/off the interactions between hidden and visible units.

\subsubsection{Hybrid Generative Nets }

Hybrid generative networks are a group of generative models that contain
both directed and undirected edges in the networks. Since studies
on hybrid approach are overshadowed by purely directed or undirected
networks, we only introduce an overview of deep belief networks (DBNs)
\citep{Hinton.Geoffrey_etal_2006NeuralCompututation_DBN}, one quintessence
of this group, that is one of the first non-convolutional deep networks
to be trained successfully. Its birth marked a historic milestone
in the development of deep learning. In the graphical perspective,
a DBN is a multi-layer neural network whose visible units lie in the
first layer and hidden units are in the remaining layers. Each unit
only connects to other units in the next upper and lower layers. The
top two layers form an RBM with undirected connections while the other
connections are directed edges pointing towards nodes in the lower
layer. As combined models, DBNs incur many problems coming from both
directed and undirected networks. For example, inference in DBNs is
difficult because of the undirected interactions between hidden units
in the top RBM and the explaining-away effects within directed layers.
Fortunately, training in $L$-layer DBNs is possible via stacking
the $L-1$ RBMs which are learned individually and layer-by-layer.
This stacking procedure is justified to guarantee to raise the lower
bound on the log-likelihood of the data \citep{Hinton.Geoffrey_etal_2006NeuralCompututation_DBN}.
After that, the network can be fine-tuned using a wake-sleep algorithm
in generative manner or its weights can be used as initialization
to learn a MLP in a discriminative fine-tuning step to perform a classification
task. Deep belief networks are developed for several applications
such as object recognition \citep{YoshuaBengio_etal_2006NIPS_objectrec,Hinton.Geoffrey_etal_2006NeuralCompututation_DBN},
classification \citep{Bengio.Yoshua_LeCun.Yann_2007_classification},
dimensionality reduction \citep{Hinton.Geoffrey_Salakhutdinov.Ruslan_2006Science}
and information retrieval \citep{Salakhutdinov.Ruslan_Hinton.Geoffrey_2007SIGIR_DBN_semantic_hashing}.
DBNs are also extended to convolutional DBNs \citep{Lee.Honglak_2009ICML_ConvDBN}
and higher-order DBNs \citep{Nair.Vinod_Hinton.Geoffrey_2009NIPS_DBN_HigherOrderBM}.

\subsection{Video Anomaly Detection}

Nowadays, due to the rise of terrorism \citep{Clarke.Melissa_2015_rise_terrorism}
and crimes \citep{CrimeStatisticasAgency_2016_rise_crime}, there
are more and more increasing concerns for security and safety in public
places and restricted areas. Due to the overabundance of surveillance
video data, extensive studies \citep{AngelaSodemann-2012-Review,Oluwatoyin-2012-Review}
have been conducted to develop intelligent systems for automatically
discovering the human behaviours in video scenes. For this reason,
in the early stage of our project, we mainly focus on building an
application of anomaly detection in surveillance systems to detect
unexpected behaviours or anomaly events in streaming videos. The anomalous
events are commonly assumed to be rare, irregular or significantly
different from the others \citep{AngelaSodemann-2012-Review}. Examples
include access to restricted area, leaving strange packages, movements
in wrong direction, fighting and falling detection, which can be captured
by the camera monitoring systems in airports, car parks, stations
and public spaces in general. Identifying abnormal behaviours allows
early intervention and in-time support to reduce the consequent cost.
Anomaly detection systems also allow to reduce the amount of data
to be processed manually by human operators via driving their attention
to a specific portion of the scenes.

The existing literature of anomaly detection in video data offers
two approaches: supervised learning and unsupervised learning. In
supervised approach, models are supplied with training data which
is annotated with normal or/and abnormal class labels. Benezeth et
al. \citep{Benezeth.Yannick_etal_2009_supervised_MarkovRandomField}
performed background subtraction to obtain motion pixels and built
a matrix that captures the co-occurrence between two motion pixels
within a spatio-temporal volume. After that, a Markov random field
whose potential function is determined by the learned co-occurrence
matrix to estimate the normality probability of observed volumes.
A Markov random field with hidden variables is also applied in \citep{Kim.Jaechul_Grauman.Kristen_2009CVPR_supervised_MPPCA}
where mixture of probabilistic principle component analysers (MPPCA)
is learned on optical flow-based features of normal image regions.
Provided a streaming scene, a Markov random field, whose parameters
are specified by the trained MPPCA, is constructed from the incoming
frame and the fixed number of recent frames. By solving the global
MAP problem, the binary values of Markov nodes are estimated to determine
normal or abnormal labels of the corresponding regions. \citet{Kratz.Louis_Nishino.Ko_2009CVPR_supervised_HMM}
computed the gradient distribution in fixed size volumes and represented
a video as a set of prototypes which are the centroids of gradient
distribution clusters. A distribution-based HMM and a coupled HMM
are trained to model the temporal and spatial correlations between
usual volumes in the crowed scenes. Social Force \citep{Mehran.Ramin_etal_2009CVPR_supervised_socialforce}
is a distinct idea, based on interaction forces between moving people
in the crowd to exact force flow representation of videos. After that,
Latent Dirichlet Allocation (LDA) can be trained on these features
to model the distribution of normal crowd behaviours. Anomaly scenes
are frames with low likelihood.

Many methods such as MPPCA mentioned \citep{Kim.Jaechul_Grauman.Kristen_2009CVPR_supervised_MPPCA}
above, chaotic invariant \citep{Wu.Shandon_etal_2010CVPR_supervised_chaotic}
and mixture of dynamic texture models (MDT) \citep{Li.WeiXin_etal_2014PAMI_supervised_MDT},
leverage on the power of mixture models to capture the normality probability
distribution. Chaotic invariant \citep{Wu.Shandon_etal_2010CVPR_supervised_chaotic}
extracts features based on maximal Lyapunov exponent and correlation
dimension to encode the chaotic information in the crowed scenes.
The probability of an observation to be normal is estimated via a
Gaussian mixture model (GMM) trained on these chaotic invariant features.
In MDT \citep{Li.WeiXin_etal_2014PAMI_supervised_MDT}, the discriminative
salience is used to measure spatial abnormality signals while the
mixture of dynamic textures are trained on normal videos to model
both appearance and temporal information. The final abnormality map
is the sum of the temporal and spatial abnormality maps. 

Several methods intend to discover the boundary of normal events from
the training data. \citet{Zhang.Ying_etal_2016PR_Supervised_MotionAppearance}
detected irregularities by integrating both motion and appearance
clues where the appearance abnormality score is the distance to the
spherical boundary, learned with support vector data description.
Sparse coding methods \citep{Lu.Cewu_etal_2013ICCV_Supervised-Sparse}
assume that the regular examples can be represented as the linear
combinations of basis vectors in a learned dictionary. Then, irregular
behaviours cause high reconstruction errors and can be distinguished
from the regularities successfully. By reformulating video anomaly
detection as classification problem, \citet{Cui.Xinyi_etal_2011CVPR_supervised_energybased}
trained SVM classifiers on combined features of interaction energy
potentials and velocities, which are extracted for every interest
point, to recognize the unusual activities. 

All supervised methods, however, require the training data annotation
which is labour-intensive for large-scale data, rendering them inapplicable
to the video streaming from surveillance systems where the amount
of data grows super-abundantly. Moreover, it is also infeasible to
model the diversity of normal event types in practice.

Unsupervised learning approach overcomes this issue by modeling the
data without the need for labels. Reconstruction-based methods, such
as principle component analysis (PCA) and sparse reconstruction, attempt
to represent the majority of data points, where the normal examples
dominate. The abnormal patterns that occur infrequently cannot be
reconstructed well by the models and cause high reconstruction errors.
PCA-based anomaly detection in \citep{Saha.Budhaditya_etal_2009ICDM_unsupervised_PCA,Pham.DucSon_etal_2011ICDM_principle_eigenvectors}
learns a linear transformation to a lower dimensional linear space
called ``residual subspace'', and then detects the anomalies using
the residual signals of the projection of this data onto the residual
subspace. Dynamic sparse coding \citep{Zhao.Bin_etal_2011CVPR_unsupervised-SparseCoding}
represents a spatio-temporal volume in the video as a set of descriptors,
e.g. HOG or HOF of interest points inside the volume. A learned compact
dictionary of basis vectors is used to reconstruct these descriptors
and identify unusual events via a measurement which evaluates the
abnormality of events. Probabilistic methods, which are able to learn
the distribution of training data, are also introduced as unsupervised
frameworks. A quintessence is GMM methods in \citep{Basharat.Arslan_etal_2008CVPR_unsupervised_GMM}.
The system in \citep{Basharat.Arslan_etal_2008CVPR_unsupervised_GMM}
firstly runs object detection and tracking modules to provide the
position information of objects over video frames. Next, GMMs are
trained on transition vectors, which are the deviation in position
at different time moments, to model regular motion patterns. The abnormal
motions can be detected due to their low likelihood values. An alternative
direction is clustering-based methods. \citet{Roshtkhari.MehrsanJavan_Levine.MartinD_2013CVPR_unsupervised_streaming_BOV}
leveraged on bag of video word models, to encode the training data
into a number of representative data points, called codewords. An
ensemble of spatio-temporal volumes is specified as abnormality if
its similarities to codewords are higher than a threshold. Scan statistics
\citep{Hu.Yang_etal_2013CVPR_unsupervised_ScanStatistics} relies
on the assumption that a normal video contains similar statistical
characteristics everywhere ($\mathcal{H}_{0}$) while a video with
abnormal regions has distinct characteristics inside the region compared
to characteristics outside ($\mathcal{H}_{1}$). Specifically, given
a spatio-temporal volume, scan statistics estimates the likelihood
ratio test statistics, assumed in the exponential form, of two above
hypotheses $\mathcal{H}_{1}$ and $\mathcal{H}_{0}$ to decide the
volume to be normal or not. \citet{Kwon.Junseok_Lee.KyoungMu_2015PAMI_unsupervised_energybased}
approached video anomaly detection problem in a distinctive way, based
on undirected graphical model. They employed a 3D segmentation algorithm
to decompose the video into a graph whose nodes are segmented regions
and edges indicate the spatio-temporal relationships between regions.
Each node is attached with meaningful characteristics such as causality
between two events $A$ and $B$ , the frequency of their co-occurrence
and the independence degree between them. By iteratively adding or
deleting edges in the graph via MCMC process, this method can learn
the optimal graph that minimizes the pre-defined energy function.
Depending on the definition of the energy function, the final graph
is able to list anomaly events or dominant events (for event summarisation
applications) in the video. For investigating temporal abnormalities,
\citet{Duong.ThiV_etal_2005CVPR_HSMM} introduced Switching Hidden
Semi-Markov Model (S-HSMM) that can estimate the probabilities of
normality over abnormality in a short period before the time $t$
to identify the abnormal durations in human activity video sequences. 

These methods, however, critically depend on the hand-crafted, low-level
features extracted for videos and images, such as gradients \citep{Roshtkhari.MehrsanJavan_Levine.MartinD_2013CVPR_unsupervised_streaming_BOV,Kwon.Junseok_Lee.KyoungMu_2015PAMI_unsupervised_energybased,Kratz.Louis_Nishino.Ko_2009CVPR_supervised_HMM,Lu.Cewu_etal_2013ICCV_Supervised-Sparse},
histograms of oriented gradients (HOG) \citep{Zhao.Bin_etal_2011CVPR_unsupervised-SparseCoding},
optical flow features \citep{Saha.Budhaditya_etal_2009ICDM_unsupervised_PCA,Hu.Yang_etal_2013CVPR_unsupervised_ScanStatistics,Kim.Jaechul_Grauman.Kristen_2009CVPR_supervised_MPPCA,Mehran.Ramin_etal_2009CVPR_supervised_socialforce,Wu.Shandon_etal_2010CVPR_supervised_chaotic,Zhang.Ying_etal_2016PR_Supervised_MotionAppearance}
and histograms of optical flow (HOF) \citep{Zhao.Bin_etal_2011CVPR_unsupervised-SparseCoding}.
This hand-crafted feature extraction relies on the design of preprocessing
pipeline and data transformation, which is labour-intensive and normally
requires exhaustive prior knowledge.

Recently there have been several studies that use deep learning techniques
to solve problems in computer vision \citep{Guo.Yanming_etal_2016NeuralComp_DL_CV_review}
in general and video anomaly detection task in particular. According
to \citep{Goodfellow.Ian_etal_2016_book}, deep learning is an approach
that is able to ``allow computers to learn from experience and understand
the world in terms of a hierarchy of concepts, with each concept defined
in terms of its relation to simpler concepts''. Due to such its capability,
deep learning is used to automatically learn high-level representation
for data to avoid the requirement of domain experts in designing features.
When applying to anomaly detection for video data, one achieves the
idea of hierarchical structure by adopting multiple-layer neural networks
or stacking shallow machine learning algorithms. 

For the first approach, one of the widely-used networks is autoencoders.
Appearance and motion deep nets (AMDNs) \citep{Xu.Dan_etal_2015BMVC_DL_StackedDenoisingAE_OCSVM}
constructed stacked denoising autoencoders on raw patches and optical
flow features to learn higher-level representation of image patches.
One-class support vector machines (OC-SVMs), which are built on top
of these autoencoders, have responsibility for estimating the abnormality
scores of events in videos. Autoencodes can also learn global representation
on cubic patches as in \citep{Sabokrou.Mohammad_etal_2015CVPRW_DL_AE_GaussianClassifier}.
These global learned features as well as local similarity between
adjacent patches, which is captured by structural similarity (SSIM)
\citep{Brunet.Dominique_etal_2012TransIP_SSIM}, are passed into Gaussian
classifiers to identify normal or abnormal patches. Both methods share
the same idea of using autoencoder networks to extract high-level
features and constructing an individual module for anomaly detection.
\citet{Hasan.Mahmudul_etla_2016CVPR_DL_ConvAE} dealed with two tasks
simultaneously via training a unique convolutional autoencoder (ConvAE)
to reconstruct the video. The reconstruction quality indicates the
abnormality degree of video frames. \citet{Hasan.Mahmudul_etla_2016CVPR_DL_ConvAE}
shows that this end-to-end framework can produce a meaningful representation
comparable with state-of-the-art handcrafted features such as HOG,
HOF and improved trajectories for anomaly detection. 

An alternative approach to implement deep architecture is to stack
machine learning algorithms together to obtain deep incremental slow
feature analysis (D-IncSFA) \citep{Hu.Xing_etal_2016ComputerVision_DL_DIncSFA}
or deep Gaussian mixture model (Deep GMM) \citep{Feng.Yachuang_etal_2017NeuralComp_DL_unsupervised_DeepGMM}.
By placing incremental slow feature analysis \citep{Kompella.VarunRaj_etal_2012NeuralComp_IncSFA}
components on top of each other, \citet{Hu.Xing_etal_2016ComputerVision_DL_DIncSFA}
constructed an end-to-end deep learning framework to both extract
features from raw data and localize anomaly events in videos. In Deep
GMMs \citep{Feng.Yachuang_etal_2017NeuralComp_DL_unsupervised_DeepGMM},
feature vectors are extracted by employing a PCANet \citet{Fang.Zhijun_etal_2016MultimediaTool_PCANet}
on the 3D gradients of image patches, and then the deep model is obtained
by layer-by-layer training GMMs to model the probability distribution
of normal patterns. Anomaly detection depends on the computation of
data likelihood to distinguish usual and unusual patches. 

Overall, most existing deep learning solutions for video anomaly detection
still partially rely on low-level features such as optical flow \citep{Xu.Dan_etal_2015BMVC_DL_StackedDenoisingAE_OCSVM},
gradients \citep{Feng.Yachuang_etal_2017NeuralComp_DL_unsupervised_DeepGMM},
and SSIM \citep{Sabokrou.Mohammad_etal_2015CVPRW_DL_AE_GaussianClassifier},
except for \citet{Hasan.Mahmudul_etla_2016CVPR_DL_ConvAE,Hu.Xing_etal_2016ComputerVision_DL_DIncSFA}
that are designed to work immediately on raw image data. In other
words, the current deep anomaly detection systems in video surveillance
have not taken the advantages of hierarchical feaure learning as the
nature of deep learning methods. This results in the need for more
intensive studies to investigate the capacity of deep learning methods
as automated frameworks in both feature representation and unsupervised
anomaly detection.

%% file: methodology.tex
The previous Section \ref{sec:literature_review} shows that deep
learning methods are rising as the next trend in anomaly detection
community in general and video anomaly detection research in particular.
Our abnormality detection system, which we are developing in this
project, is a part of this effort. In this section, we will provide
the background of RBMs, building blocks of many deep generative networks,
which we are based on to develop our anomaly detection system.

\subsection{Restricted Boltzmann Machines}

\begin{figure}[h]
\centering{}\includegraphics[width=0.5\columnwidth]{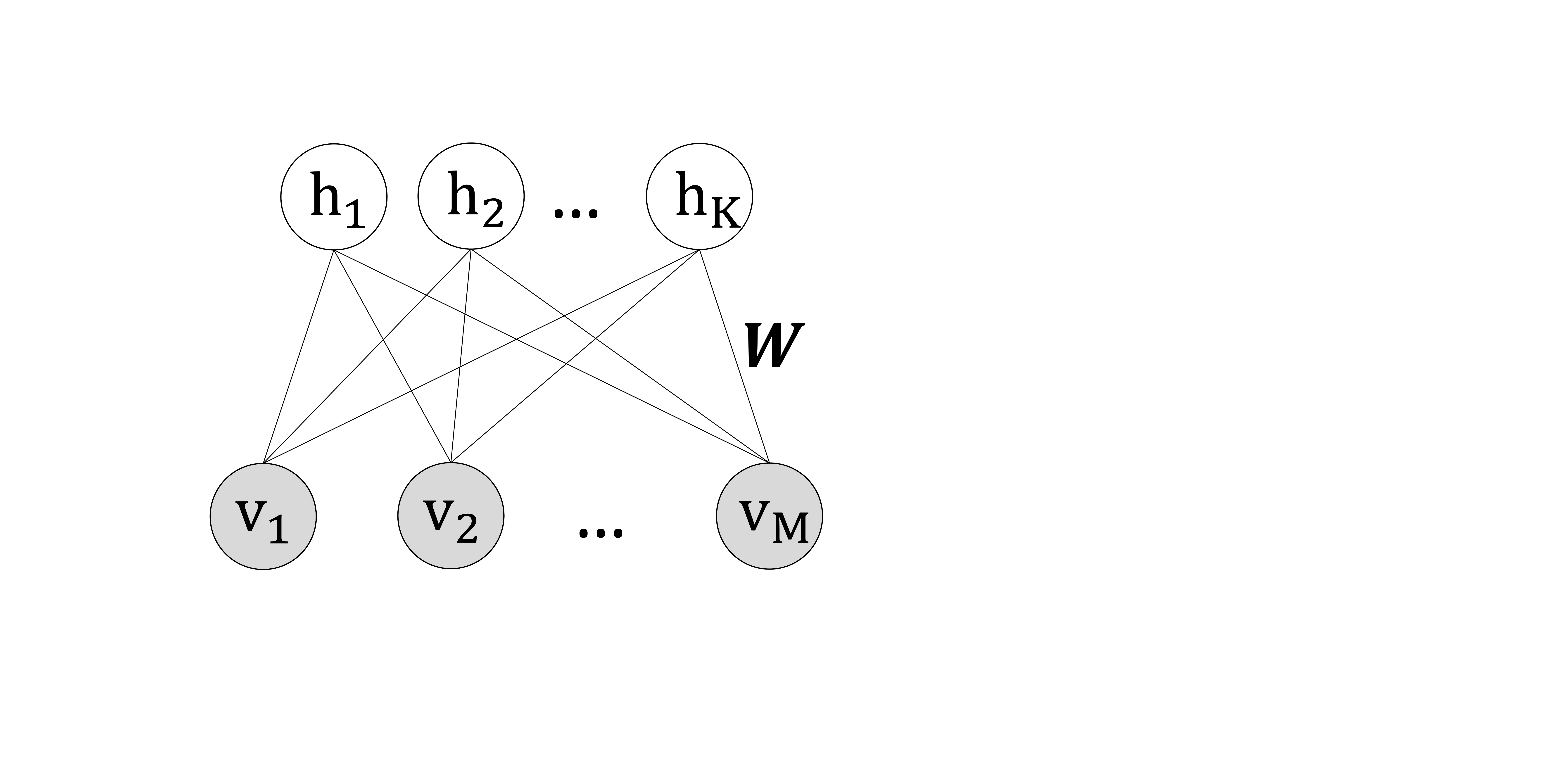}\caption{\selectlanguage{british}%
A restricted Boltzmann machine architecture of $M$ visible units
and $K$ hidden units. \selectlanguage{american}%
}
\label{fig:RBM_architecture}
\end{figure}

A restricted Boltzmann machine (RBM) \citep{Smolensky_1986,Freund_Haussler_1994_unsupervised}
is a bipartite undirected network of $M$ visible nodes and $K$ hidden
nodes. Two kinds of nodes form visible and hidden layers in RBMs.
A connection is created between any nodes in different layers but
there is no intra-layer connection. The ``restriction'' on internal
connections in layers brings RBM models the efficiency in learning
and inference which cannot be obtained in general Boltzmann machines
\citep{Ackley.DavidH_etal_1985_BM}. The architecture of an RBM is
shown in Fig. \ref{fig:RBM_architecture}. Each node in the graph
is associated with a random variable whose type can be binary \citep{Welling.Max_etal_2005NIPS_exponential},
integer \citep{Welling.Max_etal_2005NIPS_exponential}, continuous
\citep{Hinton.Geoffrey_etal_2006NeuralCompututation_DBN} or categorical,
depending on particular data types. In the first year of HDR candidature,
we only focus on binary restricted Boltzmann machines and their details
will be introduced in this section.

\paragraph{Model representation}

Let use consider an RBM with $M$ visible variables $\bv=\lrc{v_{1},v_{2},\ldots,v_{M}}\in\lrc{0,1}^{M}$
and $K$ latent variables $\bh=\{h_{1},h_{2},...,h_{K}\}\in\{0,1\}^{K}$.
The model is parameterized by two bias vectors $\ba=\{a_{1},a_{2},...,a_{M}\}\in\mathbb{R}^{M}$
in the visible layer and $\boldsymbol{b}=\{b_{1},b_{2},...,b_{M}\}\in\mathbb{R}^{K}$
in the hidden layer and a weight matrix $\boldsymbol{W}\in\mathbb{R}^{M\times K}$
whose element $w_{ij}$ is the weight of the edge from the visible
unit $v_{i}$ and the hidden unit $h_{j}$. The set $\boldsymbol{\psi}=\{\boldsymbol{a},\boldsymbol{b},\mathbf{W\}}$
describes the network parameters and specifies RBM's power. 

Similar to Boltzmann machines, an RBM is an energy-based model and
its energy value, defined at a particular value of a pair $\bv$ and
$\bh$, is given by: 
\begin{align}
E(\bv,\boldsymbol{h};\boldsymbol{\psi}) & =-(\mathbf{a^{\top}}\bv+\boldsymbol{b}^{\top}\boldsymbol{h}+\bv^{\top}\bW)\label{eq:RBM_energy_def}
\end{align}

Following Boltzmann distribution, as known as Gibbs distribution,
the joint probability is defined via the energy function as: 
\begin{align}
p(\bv,\boldsymbol{h};\boldsymbol{\psi}) & =\frac{1}{\partfunc(\boldsymbol{\psi})}e^{-E(\bv,\boldsymbol{h};\boldsymbol{\psi})}\label{eq:RBM_pvh_def}
\end{align}
where, $\partfunc(\boldsymbol{\psi})$ is the \textit{partition function}
which is the sum of $e^{-E(\bv,\boldsymbol{h};\boldsymbol{\psi})}$
over all possible pairwise configurations of visible and hidden units
in the network, and therefore it guarantees that $p(\bv,\boldsymbol{h};\boldsymbol{\psi})$
is a proper distribution. 
\begin{equation}
\partfunc\lrr{\bpsi}=\sum_{\bv,\bh}e^{-E(\bv,\boldsymbol{h};\boldsymbol{\psi})}\label{eq:RBM_partfunc_def}
\end{equation}

By marginalizing Eq. \ref{eq:RBM_pvh_def} out latent variables $\bh$,
we obtain the data likelihood which assigns a probability to each
observed data $\bv$:
\begin{eqnarray}
p(\bv;\psi) & = & \frac{1}{\partfunc(\psi)}\sum_{\boldsymbol{h}}e^{-E(\bv,\boldsymbol{h};\psi)}\label{eq:RBM_likelihood}
\end{eqnarray}

Thanks to the special bipartite structure, RBMs offer a convenient
way to compute the conditional probabilities $p(\bv|\boldsymbol{h};\boldsymbol{\psi})$
and $p(\boldsymbol{h}|\bv;\boldsymbol{\psi})$. Indeed, since units
in a layer are conditionally independent of units in the other layer,
the conditional probabilities can be nicely factorized as
\begin{align}
p(\bv|\boldsymbol{h};\boldsymbol{\psi}) & =\prod_{i=1}^{M}p(v_{i}|\boldsymbol{h};\boldsymbol{\psi})\qquad p(v_{i}=1|\bh,\boldsymbol{\psi})=\sigma(a_{i}+\mathbf{w}_{i\cdotp}\bh)\label{eq:visible_factorization}\\
p(\boldsymbol{h}|\bv;\boldsymbol{\psi}) & =\prod_{j=1}^{K}p(h_{j}|\bv;\boldsymbol{\psi})\qquad p(h_{j}=1|\bv,\boldsymbol{\psi})=\sigma(b_{j}+\bv\mathbf{^{\top}w}_{\cdotp j})\label{eq:hidden_fractorization}
\end{align}
where $\sigma\lrr x=1/\lrr{1+e^{-x}}$ is a logistic function. In
other words, the factorial property allows us to represent the conditional
distribution over a layer as the product of the distributions over
individual random variables in the layer.

\paragraph{Parameter estimation}

As a member of energy-based models, the RBM training aims to search
for a parameter set $\hat{\bpsi}$ that minimizes the network energy.
According to the inverse proportion between the data likelihood and
the energy function in Eq. \ref{eq:RBM_likelihood}, the optimization
is restated as maximizing the log-likelihood of observed data $\bv$:
\begin{equation}
\log\mathcal{L}\lrr{\bv}=\log p(\bv;\boldsymbol{\psi})=\sum_{\bh}p(\bv,\bh;\boldsymbol{\psi})\label{eq:RBM_loglikelihood}
\end{equation}

Unfortunately, since the maximal points of $\log\LL\lrr{\bv}$ cannot
be expressed in the closed-form, a gradient ascent procedure is usually
used to iteratively update the parameters in the gradient direction
\begin{align}
\boldsymbol{\psi} & =\boldsymbol{\psi}+\eta\grad\log\mathcal{L}\lrr{\bv}\label{eq:RBM_parameter_update}
\end{align}
wherein, $\eta$ is a pre-defined learning rate and the gradient vector
$\grad\log\mathcal{L}\lrr{\bv}$ is computed as follows: 
\begin{align}
\frac{\partial\log\mathcal{L}(\bv;\boldsymbol{\psi})}{\partial\boldsymbol{\psi}} & =\mathbb{E}_{p\left(\boldsymbol{h}|\bv;\boldsymbol{\psi}\right)}[-\frac{\partial E(\bv,\boldsymbol{h};\boldsymbol{\psi})}{\partial\boldsymbol{\psi}}]-\mathbb{E}_{p\left(\bv,\boldsymbol{h};\boldsymbol{\psi}\right)}[-\frac{\partial E(\bv,\boldsymbol{h};\boldsymbol{\psi})}{\partial\boldsymbol{\psi}}]\label{eq:RBM_gradient}
\end{align}

\begin{figure}[h]
\centering{}\includegraphics[width=0.8\columnwidth]{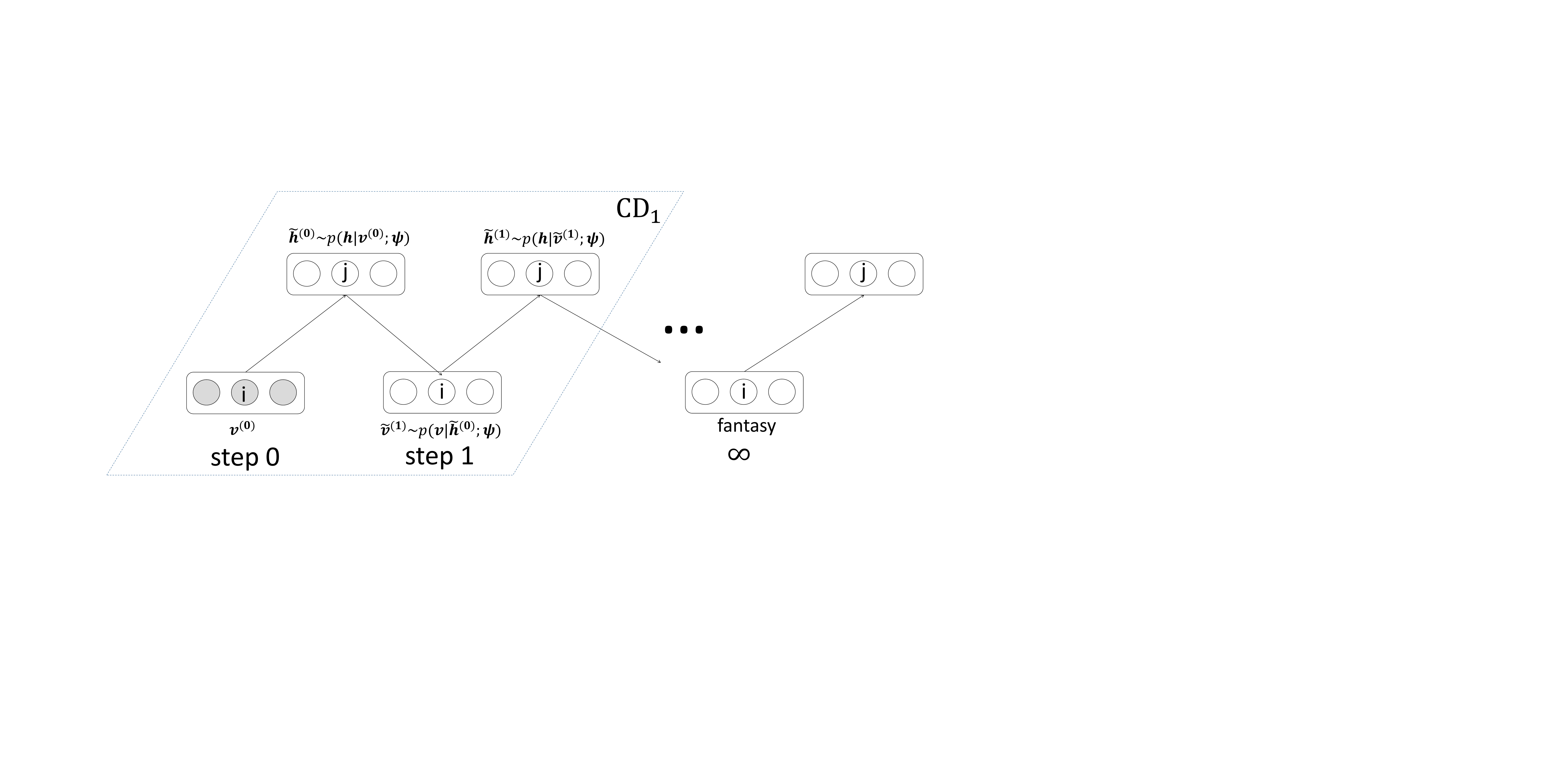}\caption{\selectlanguage{british}%
An illustration of Gibbs sampling and 1-step contrastive divergence.\selectlanguage{american}%
}
\label{fig:RBM_Gibbs}
\end{figure}

The first term on the right-hand side, denoted $\mathbb{E}_{p\left(\boldsymbol{h}|\bv;\boldsymbol{\psi}\right)}$,
 is called the \emph{data expectation} (also known as the positive
phase) that is the expected value over the posterior distribution
of given data $\bv$ whilst the second expectation, $\mathbb{E}_{p\left(\bv,\boldsymbol{h};\boldsymbol{\psi}\right)}$,
is known as the \emph{model expectation }(or the negative phase) that
is the statistics over the model distribution defined in Eq. \ref{eq:RBM_pvh_def}.
For general Boltzmann machines, both terms are intractable, and hence
it is not easy to learn a Boltzmann machine. Interestingly, by factorizing
conditional probability $p\lrr{\bh|\bv;\bpsi},$ RBMs can compute
the data expectation analytically. However, the second expectation
evaluation is still a challenging problem. To overcome this issue,
Markov chain Monte Carlo is widely-used to approximate such expectation.
More especially, the capability of conditional probability factorization
in RBMs enables us to do sampling efficiently using Gibbs sampling
\citep{Geman.Stuart_Geman.Donald_1984_GibbsSampling}. Intuitively,
we alternatively draw hidden and visible samples from the conditional
probability distributions given the other variables (Eqs. \ref{eq:visible_factorization}
and \ref{eq:hidden_fractorization}), $\tilde{\bh}\sim p(\bh|\bv;\bpsi)$
and $\tilde{\bv}\sim p(\bv|\bh;\bpsi)$, in one Gibbs sampling step.
To gain the unbiased estimate of the gradient, it is necessary for
the Markov chain to converge to the equilibrium distribution. By this
way, the training is viewed as adjusting the parameters to minimize
the Kullback-Lieber divergence between the data distribution and the
equilibrium distribution. To speed up the approximation process, Hinton
proposed to use contrastive divergence with $m$ Gibbs sampling steps
(denoted $\text{CD}_{m}$) to evaluate the model expectation. He argued
that when the Markov chain converges, the distributions between two
consecutive sampling steps are almost the same. As a result, $\text{CD}_{m}$
minimizes the difference between the data distribution and the $m$-sampling
step distribution rather than equilibrium distribution. Although contrastive
divergence is fast and has low variance, it is still far from the
equilibrium distribution if the mixing rate is low. An alternative
is persistent contrastive divergence (PCD)\citep{Tieleman.Tijmen_2008ICML_PCD}.
Unlike CD whose MCMC chain is restarted for each data sample, PCD's
chain is only reset after a regular interval or not reset at all.
Furthermore, PCD maintains several chains, usually equal to the batch
size, at the same time to achieve a better approximation. Fig. \ref{fig:RBM_Gibbs}
demonstrates the alternative steps of Gibbs sampler, where $\CDm$
is considered as its truncated version with the first $m+1$ steps.
$\CDone$ is common in practice since it shows a large improvement
in training time with only a small bias \citep{MiguelCarreiraPerpinan_etal_2005_CD}.

\paragraph{Data reconstruction}

Once the optimal parameters $\hat{\bpsi}$ are learned, the model
can be used to obtain the reconstructed data $\tilde{\bv}$ whenever
supplying an input data $\bv$. Firstly, by propagating the input
vector to the hidden layer, we acquire its new representation $\tilde{\bh}=\lrr{\tilde{h}_{1},\tilde{h}_{2},...,\tilde{h}_{K}}^{\top}$
in the hidden space as follows:
\begin{equation}
\tilde{h}_{k}\sim p\left(h_{k}=1\gv\bv\right)=\sigma\left(b_{k}+\sum_{m}w_{mk}v_{m}\right)\label{eq:RBM_forward_and_sampling}
\end{equation}

The hidden sample vector $\tilde{\bh}$ is then mapped back to the
input space for the reconstructed data $\tilde{\bv}=\left[\tilde{v}_{1},\tilde{v}_{2},...,\tilde{v}_{M}\right]^{\top}$
where
\begin{equation}
\tilde{v}_{m}\sim p\left(v_{m}=1\gv\tilde{\bh};\psi\right)=\sigma\left(a_{m}+\sum_{k}w_{mk}\tilde{h}_{k}\right)\label{eq:RBM_back_and_sampling}
\end{equation}

The forward and backward propagations in Eqs. \ref{eq:RBM_forward_and_sampling}
and \ref{eq:RBM_back_and_sampling} are done effectively because of
the nice factorisation nature of RBMs. More practically, the reconstructed
data can be used to recover the missing or corrupted elements in $\bv$
(caused by sensory errors or transmission noises) or do classification
\citep{Larochelle.Hugo_Bengio.Yoshua_2008ICML_discriminativeRBM}
while the high difference between $\bv$ and $\dot{\bv}$ indicates
an occurrence of anomaly signals or data.

\subsection{Framework}

We propose a unified framework for anomaly detection in video based
on the restricted Boltzmann machine ($\model$) \citep{Freund_Haussler_1994_unsupervised,Hinton.Geoffrey_2002NeuralComp_CD}.
Our proposed system employs $\model$s as core modules to model the
complex distribution of data, capture the data regularity and variations
\citep{Nguyen.TuDinh_etal_2015AAAI_TvRBM}, as a result effectively
reconstruct the normal events that occur frequently in the data. The
idea is to use the errors of reconstructed data to recognize the abnormal
objects or behaviours that deviate significantly from the common. 

Our framework is trained in a completely unsupervised manner that
does not involve any explicit labels or implicit knowledge of what
to be defined as abnormal. In addition, it can work directly on raw
pixels without the need for expensive feature engineering procedure.
Another advantage of our method is the capability of detecting the
exact boundary of local abnormality in video frames. To handle the
video data coming in a stream, we further extend our method to incrementally
update parameters without retraining the models from scratch. Our
solution can be easily deployed in arbitrary surveillance streaming
setting without the expensive calibration requirement.

\begin{figure}
\begin{centering}
\emph{\includegraphics[width=1\columnwidth]{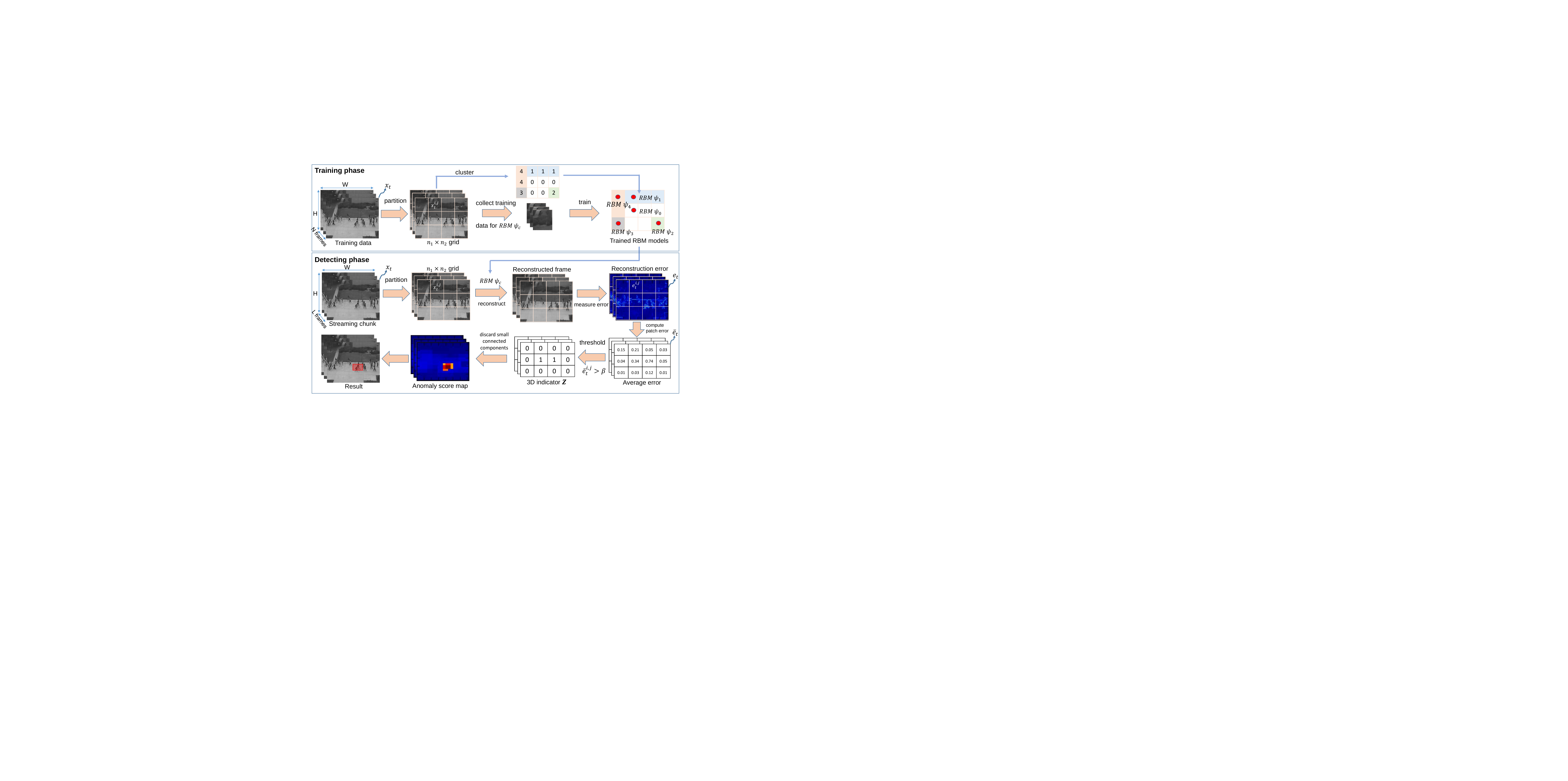}}
\par\end{centering}

\caption{The overview of our proposed framework.\label{figFramework}}
\end{figure}

We now describe our proposed framework that is based on the $\model$
to detect anomaly events for each frame in video data. In general,
our system is a two-phase pipeline: training phase and detecting phase.
Particularly in the training phase, our model: (i) takes a series
of video frames in the training data as a collection of images, (ii)
divides each image into patches, (iii) gathers similar patches into
clusters, and (iv) learns separate $\model$ for each cluster using
the image patches. The detecting phase consists of three steps: (i)
collecting image patches in the testing video for each cluster, and
then using the learned $\model$ to reconstruct the data for the corresponding
cluster of patches, (ii) proposing the regions that are \emph{potential}
to be abnormal by applying a predefined threshold to reconstruction
errors, and then finding connected components of these candidates
and filtering out those too small, and (iii) updating the model incrementally
for the data stream. The overview of our framework is illustrated
in Fig.~\ref{figFramework}. In what follows, we describe training
and detecting phases in more details.

\paragraph{Training phase.}

Assume that the training data consists of $N$ video frames with the
size of $H\times W$ pixels, let denote $\mathcal{D}=\{\bx_{t}\in\mathbb{R}^{H\times W}\}_{t=1}^{N}$.
In real-life video surveillance data, $H\times W$ is usually very
large (e.g., hundreds of thousand pixels), hence it is often infeasible
for a single $\model$ to handle such high-dimensional image. This
is because the high-dimensional input requires a more complex model
with an extremely large number of parameters (i.e., millions). This
makes the parameter learning more difficult and less robust since
it is hard to control the bounding of hidden activation values. Thus
the hidden posteriors are easily collapsed into either zeros or ones,
and no more learning occurs.

To tackle this issue, one can reduce the data dimension using dimensionality
reduction techniques or by subsampling the image to smaller size.
This solution, however, is computational demanding and may lose much
information of the original data. In this work we choose to apply
$\model$s directly to raw imaginary pixels whilst try to preserve
information. To that end, we train our model on $h\times w$ patches
where we divide each image $\bx_{t}$ into a grid of $N_{h}\times N_{w}$
patches: $\bx_{t}=\{\bx_{t}^{i,j}\gv1\leq i\leq N_{h},1\leq j\leq N_{w}\}$.
This approach greatly reduces the data dimensionality and hence requires
smaller models. One way is to learn independent $\model$s on patches
at each location ($i,j$). However, this would result in an excessive
number of models, for example, $400$ $\model$s to work on the $240\times360$
image resolution and $12\times18$ patch size, hence leading to very
high computational complexity and memory demand.

Our solution is to reduce the number of models by grouping all similar
patches from different locations for learning a single model. We observe
that it is redundant to train a separate model for each location of
patches since most adjacent patches such as pathways, walls and nature
strips in surveillance scenes have similar appearance and texture.
Thus we first train an\textit{ $\model$} with a small number of hidden
units ($K=4$) on all patches $\{\bx_{t}^{i,j}\}$ of all video frames.
We then compute the hidden posterior $\tilde{\bh}$ for each image
patch $\bx_{t}^{i,j}$ and binarize it to obtain the binary vector:
$\tilde{\bh}=\left[\mathbb{I}\left(\tilde{h}_{1}>0.5\right),...,\mathbb{I}\left(\tilde{h}_{K}>0.5\right)\right]$
where $\mathbb{I}\left(\bullet\right)$ is the indicator function.
Next this binary vector is converted to an integer value in decimal
system, e.g., 0101 converted to 5, which we use as the \emph{pseudo-label}
$\lambda_{t}^{i,j}$ of the cluster of the image patch $\bx_{t}^{i,j}$.
The cluster label $c^{i,j}$ for all patches at location ($i,j$)
is chosen by voting the pseudo-labels over all $N$ frames: $\lambda_{1}^{i,j},\lambda_{2}^{i,j},...,\lambda_{N}^{i,j}$.
Let $C$ denote the number of unique cluster labels in the set \{$c^{i,j}\gv1\leq i\leq N_{h},1\leq j\leq N_{w}$\},
we finally train $C$ independent\textit{\emph{ $\model$s}}\emph{
}with a larger number of hidden units ($K=100$), each with parameter
set $\psi_{c}$ for all patches with the same cluster label $c$.

\paragraph{Detecting phase.}

Once all $\model$s have been learned using the training data, they
are used to reveal the irregular events in the testing data. The pseudocode
of this phase is given in Alg.~\ref{algDetectionPhase}. Overall,
there are three main steps: reconstructing the data, detecting local
abnormal objects and updating models incrementally. In particular,
the stream of video data is first split into chunks of $L$ non-overlapping
frames, each denoted by $\left\{ \bx_{t}\right\} _{t=1}^{L}$. Each
patch $\bx_{t}^{i,j}$ is then reconstructed to obtain the reconstruction
$\tilde{\bx}_{t}^{i,j}$ using the learned $\model$ with parameters
$\psi_{c^{i,j}}$, and all together form the reconstructed data $\tilde{\bx}_{t}$
of the frame $\bx_{t}$. The reconstruction error $\be_{t}=[\be_{t}^{i,j}]\in\realset^{H\times W}$
is then computed as: $\be_{t}^{i,j}=|\bx_{t}^{i,j}-\tilde{\bx}_{t}^{i,j}|$.

To detect abnormal pixels, one can compare the reconstruction error
$\be_{t}$ with a given threshold. This approach, however, may produce
many false alarms when normal pixels are reconstructed with high errors,
and may fail to cover the entire abnormal objects in such a case that
they are fragmented into isolated high error parts. Our solution is
to work on the average error $\bar{e}_{t}^{i,j}=||\boldsymbol{e}_{t}^{i,j}||_{2}/\left(h\times w\right)$
over patches rather than individual pixels. These errors are then
compared with a predefined threshold $\beta$. All pixels in the patch
$\bx_{t}^{i,j}$ are considered abnormal if $\bar{e}_{t}^{i,j}\geq\beta$.

Applying the above procedure and then concatenating $L$ frames, we
obtain a binary 3D rectangle $\bZ\in\left\{ 0,1\right\} ^{L\times H\times W}$
wherein $z_{i,j,k}=1$ indicates the abnormal voxel whilst $z_{i,j,k}=0$
the normal one. Throughout the experiments, we observe that most of
abnormal voxels in $\bZ$ are detected correctly, but there still
exist several small groups of voxels are incorrect. We further filter
out these false positive voxels by connecting all their \emph{related}
neighbors. More specifically, we first build a sparse graph whose
nodes are abnormal voxels $z_{i,j,k}=1$ and edges are the connections
of these voxels with their abnormal neighbours $z_{i+u,j+v,k+t}=1$
where $u,v,t\in\left\{ -1,0,1\right\} $ and $\left|u\right|+\left|v\right|+\left|t\right|>0$.
We then find all connected components in this graph, and discard small
components spanning less than $\gamma$ contiguous frames. The average
error $\bar{e}_{t}^{i,j}$ after this component filtering step can
be used as final anomaly score. 


\begin{algorithm}[H] 
\begin{algorithmic}[1]
\Require{ Video chunk $\left\{ \bx_{t}\right\} _{t=1}^{L}$, models $\left\{ \psi_{c}\right\} _{c=1}^{C}$, thresholds $\beta$ and $\gamma$}
\Ensure{Detection $\boldsymbol{Z}$, score $\left\{ \bar{e}_{t}^{i,j}\right\} $}
\For {$t\leftarrow1,\ldots,L$}
\For {$\boldsymbol{x}_{t}^{i,j}\in\boldsymbol{x}_{t}$}
\State{$\tilde{\bx}_{t}^{i,j}\leftarrow$reconstruct($\bx_{t}^{i,j}$,$\psi_{c^{i,j}}$)}
\State{$\boldsymbol{e}_{t}^{i,j}\leftarrow|\bx_{t}^{i,j}-\tilde{\bx}_{t}^{i,j}|$}
\State{$\bar{e}_{t}^{i,j}\leftarrow\frac{1}{h\times w}\left\Vert \boldsymbol{e}_{t}^{i,j}\right\Vert _{2}$}
\If{$\bar{e}_{t}^{i,j}\geq\beta$} 
\For {$p\in\boldsymbol{x}_{t}^{i,j}$}
\State{$\bZ(p)\leftarrow1$}
\EndFor
\Else
\For {$p\in\boldsymbol{x}_{t}^{i,j}$}
\State{$\bZ(p)\leftarrow0$}
\EndFor
\EndIf
\EndFor
\For {$c\leftarrow1,\ldots,C$}
\State{$\boldsymbol{X}_{t}^{c}\leftarrow\left\{ \boldsymbol{x}_{t}^{i,j}\mid c^{i,j}=c\right\} $}
\State{$\psi_{c}\leftarrow\text{{updateRBM}}(\boldsymbol{X}_{t}^{c},\psi_{c})$}
\EndFor
\EndFor
\State{$\boldsymbol{Z}\leftarrow$remove\_small\_components($\boldsymbol{Z},$$\gamma$)}
\end{algorithmic}

\caption{RBM anomaly detection} 
\label{algDetectionPhase} 
\end{algorithm}

In the scenario of streaming videos, the scene frequently changes
over time and it could be significantly different from those are used
to train $\model$s. To tackle this issue, we further extend our proposed
framework to enable the $\model$s to adapt themselves to the new
video frames. For every incoming frame $t$, we extract the image
patches and update the parameters $\psi_{1:C}$ of $C$ $\model$s
in our framework following the procedure in the training phase. Recall
that the $\model$ parameters are updated iteratively using gradient
ascent, thus here we use several epochs to ensure the information
of new data are sufficiently captured by the models.

One problem is the anomalous objects can be presented in different
sizes in the video. To deal with this issue, we apply our framework
to the video data at different scales whilst keeping the same patch
size $h\times w$. This would help the patch partially or entirely
cover objects at certain scales. To that end, we rescale the original
video into different resolutions, then employ the same procedure above
to compute the average reconstruction error map $\bar{e}_{t}$ and
3D rectangular indicators $\bZ$. The average error maps are then
aggregated into one matrix using max operation. Likewise, indicator
tensors are merged into one before finding the connected components.
We also use overlapping patches to localize anomalous objects more
accurately. Pixels in the overlapping regions are averaged when combining
patches into the whole map.

%% file: results.tex
In this section, we empirically evaluate the performance of our anomaly
detection framework both qualitatively and quantitatively. Our aim
is to investigate the capabilities of capturing data regularity, reconstructing
the data and detecting local abnormalities of our system. For quantitative
analysis, we compare our proposed method with several up-to-date baselines.

We use 3 public datasets: UCSD Ped 1, Ped 2 \citep{Li.WeiXin_etal_2014PAMI_supervised_MDT}
and Avenue \citep{Lu.Cewu_etal_2013ICCV_Supervised-Sparse}. Under
the unsupervised setting, we disregard labels in the training videos
and train all methods on these videos. The learned models are then
evaluated on the testing videos by computing 2 measures: area under
ROC curve (AUC) and equal error rate (EER) at frame-level (no anomaly
object localization evaluation) and pixel-level (40\% of ground-truth
anomaly pixels are covered by detection), following the evaluation
protocol used in \citep{Li.WeiXin_etal_2014PAMI_supervised_MDT} and
at dual-pixel level (pixel-level constraint above and at least $\alpha$
percent of detection is true anomaly pixels) in \citep{Sabokrou.Mohammad_etal_2015CVPRW_DL_AE_GaussianClassifier}.
Note that pixel-level is a special case of dual-pixel where $\alpha=0$.
Since the videos are provided at different resolution, we first resize
all into the same size of $240\times360$.

For our framework, we duplicate and rescale video frames to multi-scale
copies with the ratios of $1.0$, $0.5$ and $0.25$, and then use
$12\times18$ image patches with $50\%$ overlapping between two adjacent
patches. Each $\model$ now consists of $216$ visible units and $4$
hidden units for clustering step whilst $100$ hidden units for training
and detecting phases. All RBMs are trained using $\text{\ensuremath{\text{CD}_{1}}}$
with learning rate $\eta=0.1$. To simulate the streaming setting,
we split testing videos in non-overlapping chunks of $L=20$ contiguous
frames and use $20$ epochs to incrementally update parameters of
$\model$s. The thresholds $\beta$ and $\gamma$ to determine anomaly
are set to $0.003$ and $10$ respectively. Those hyperparameters
have been tuned to reduce false alarms and to achieve the best balanced
AUC and EER scores.

\subsubsection{Region Clustering}

\begin{figure}[t]
\begin{centering}
\includegraphics[width=1\columnwidth]{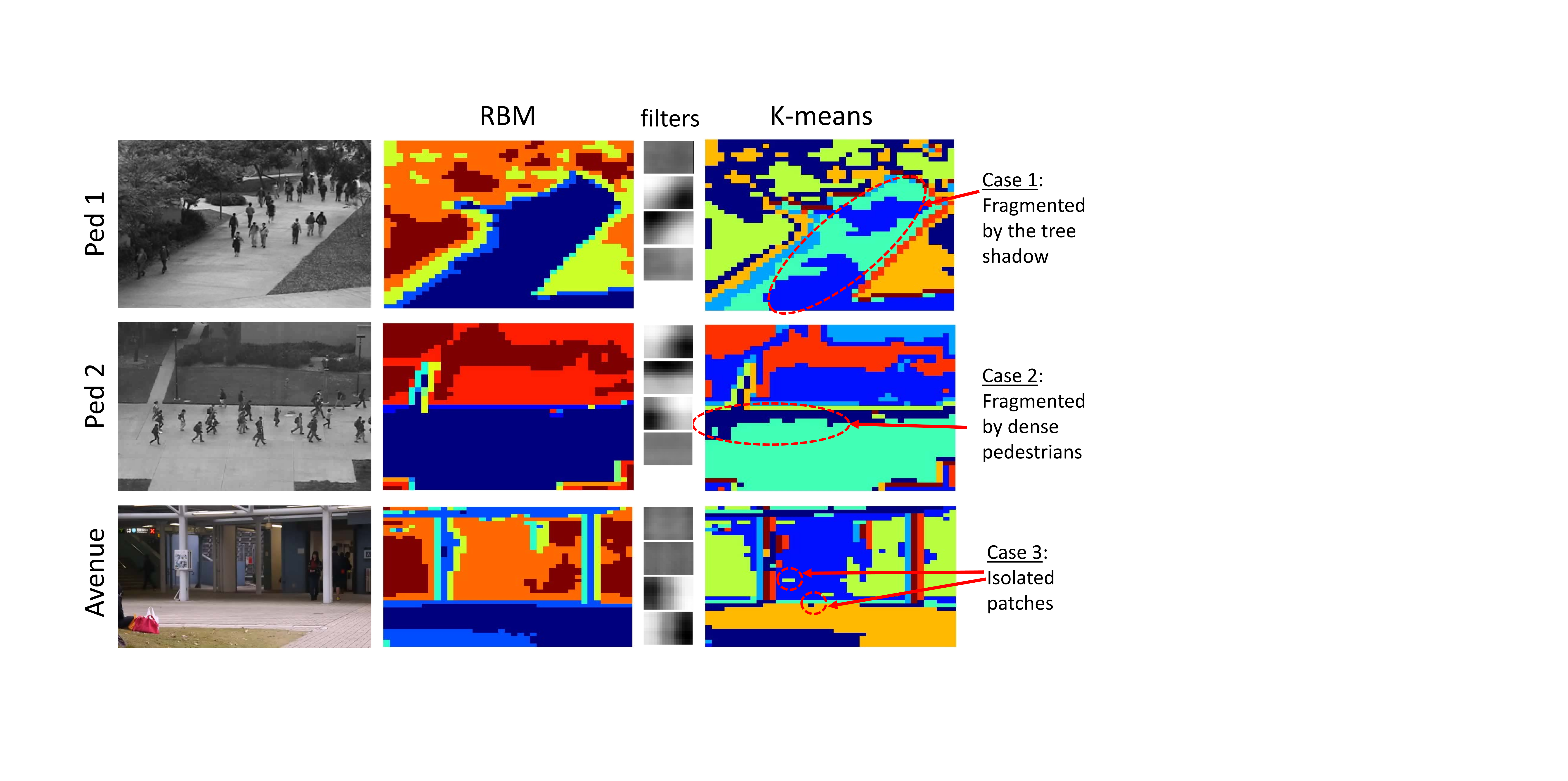}\vspace{-1em}

\par\end{centering}

\caption{Clustering results on some surveillance scenes at the first scale:
(first column) example frames; (second) cluster maps produced by $\protect\model$;
(third) filters learned by $\protect\model$; and (fourth) cluster
maps produced by $k$-means.\label{figCluster}}
\vspace{-2em}
\end{figure}

In the first experiment, we examine the clustering performance of
$\model$. Fig.~\ref{figCluster} shows the cluster maps discovered
by RBM on three datasets. Using $4$ hidden units, the $\model$ can
produce a maximum of $16$ clusters, but in fact, the model returns
less and varied number of clusters for different datasets at different
scales. For example, ($6$, $7$, $10$) similar regions at scales
($1.0$, $0.5$, $0.25$) are found for Ped 1 dataset, whilst these
numbers for Ped 2 and Avenue dataset are ($9$, $9$, $8$) and ($6$,
$9$, $9$) respectively. This suggests the capability of automatically
selecting the appropriate number of clusters of $\model$.

For comparison, we run $k$-means algorithm with $k=8$ clusters,
the average number of clusters of RBM. It can be seen from Fig.\ \ref{figCluster}
that the $k$-means fails to connect large regions which are fragmented
by the surrounding and dynamic objects, for example, the shadow of
tree on the footpath (Case 1), pedestrians walking at the upper side
of the footpath (Case 2). It also assigns several wrong labels to
small patches inside a larger area as shown in Case 3. By contrast,
the $\model$ is more robust to the influence of environmental factors
and dynamic foreground objects, and thus produces more accurate clustering
results. Taking a closer look at the filters learned by $\model$
at the third column in the figure, we can agree that the $\model$
learns the basic features such as homogeneous regions, vertical, horizontal,
diagonal edges and corners, which then can be combined to construct
the entire scene.

\subsubsection{Data Reconstruction}

\begin{figure}[h]
\centering{}\includegraphics[width=1\columnwidth]{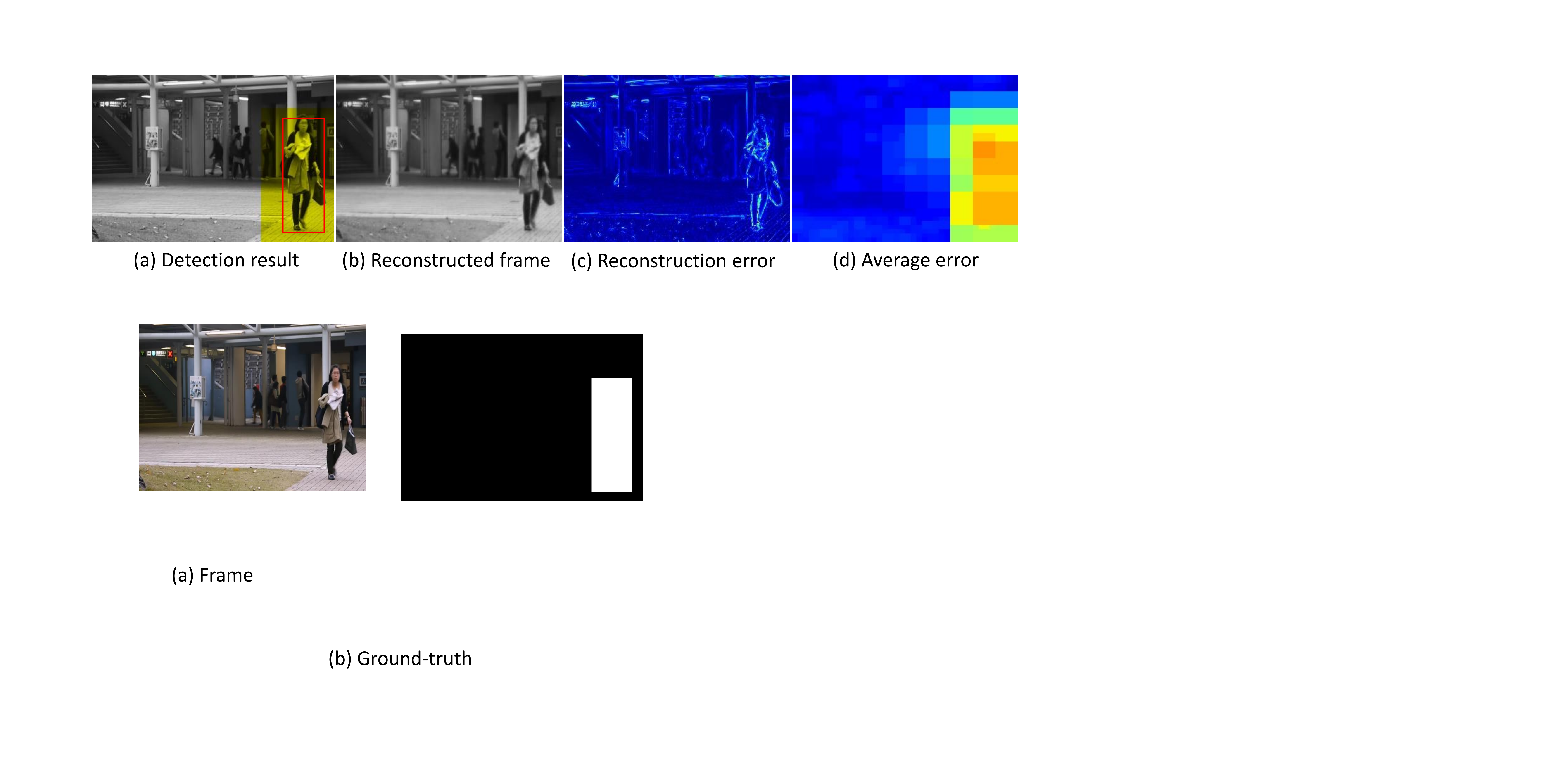}\vspace{-1em}
\caption{Data reconstruction of our method on Avenue dataset: (a) the original
frame with detected outlier female (yellow region) and ground-truth
(red rectangle), (b) reconstructed frame, (c) reconstruction error
image, (d) average reconstruction errors of patches.\label{figExample}}
\end{figure}

We next demonstrate the capability of our framework on the data reconstruction.
Fig.~\ref{figExample} shows an example of reconstructing the video
frame in Avenue dataset. Here the abnormal object is a girl walking
toward the camera. It can be seen that our model can correctly locate
this outlier behaviour based on the reconstruction errors shown in
figures (c) and (d). This is because the $\model$ can capture the
data regularity, thus produces low reconstruction errors for regular
objects and high errors for irregular or anomalous ones as shown in
figures (b) and (c).

\begin{figure}[h]
\centering{}\includegraphics[width=1\columnwidth]{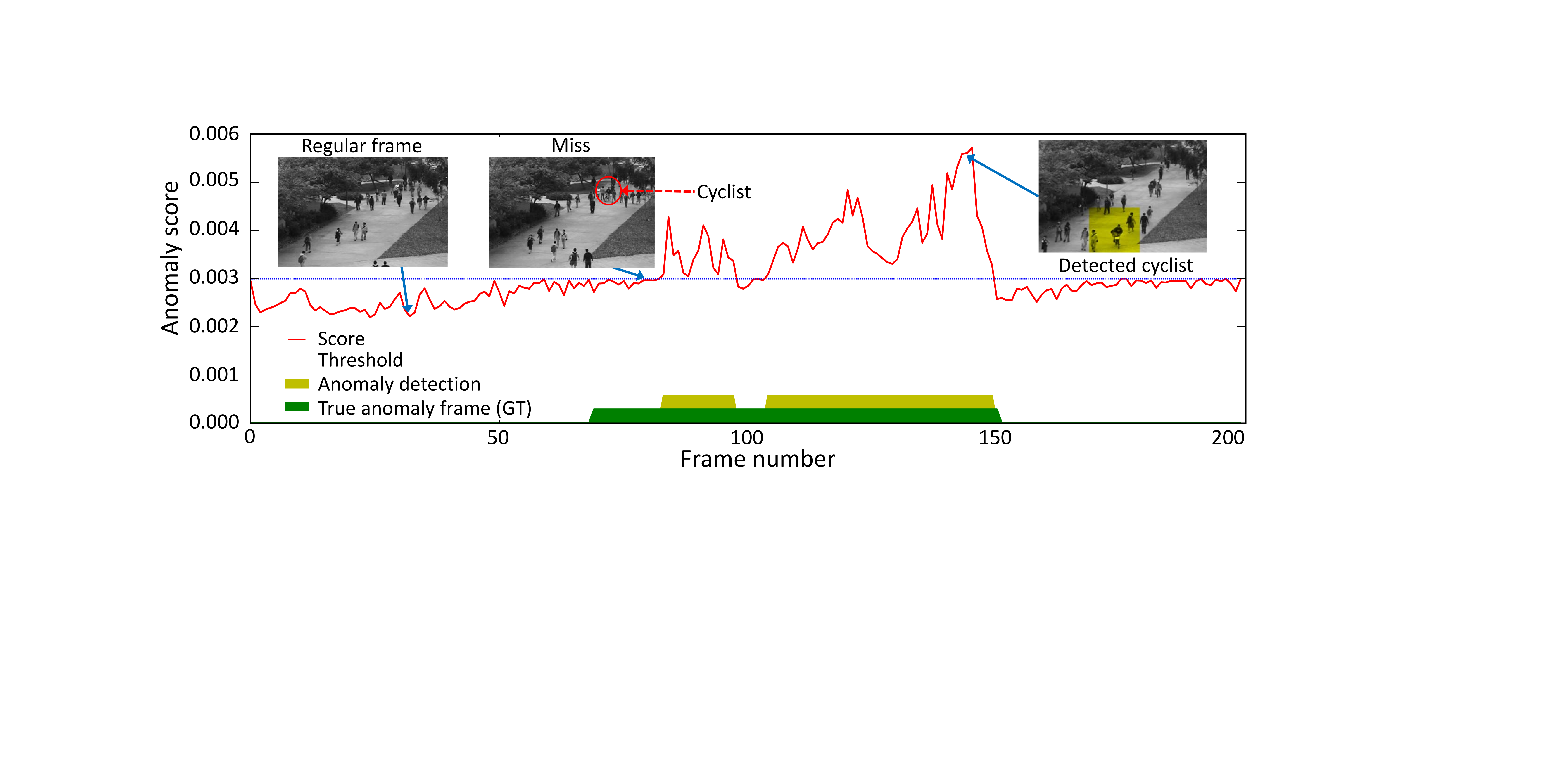}\vspace{-1em}
\caption{Average reconstruction error per frame in test video \#1 of UCSD Ped
1 dataset. The shaded green region illustrates anomalous frames in
the ground truth, while the yellow anomalous frames detected by our
method. The blue line shows the threshold.\label{figAnomScore}}
\vspace{-1em}
\end{figure}

To examine the change of reconstruction errors in a stream of video
frames, we visualize the maximum average reconstruction error in a
frame as a function of frame index as shown in Fig.~\ref{figAnomScore}.
The test video \#1 in UCSD Ped 1 dataset contains some normal frames
of walking on a footpath, followed by the appearance of a cyclist
moving towards the camera. Our system could not detect the emergence
of the cyclist since the object is too small and cluttered by many
surrounding pedestrians. However, after several frames, the cyclist
is properly spotted by our system with the reconstruction errors far
higher than the threshold.

\subsubsection{Anomaly Detection Performance }

In the last investigation, we compare our offline RBM framework and
its streaming version (called S-RBM) with the unsupervised methods
for anomaly detection in the literature. We use 4 baselines for comparison:
principal component analysis (PCA), one-class support vector machine
(OC-SVM) \citep{Amer.Mennatallah_etal_2013SIGKDD_OCSVM}, Gaussian
mixture models (GMM), and convolutional autoencoder (ConvAE) \citep{Hasan.Mahmudul_etla_2016CVPR_DL_ConvAE}.
We use the variant of PCA with optical flow features from \citep{Saha.Budhaditya_etal_2009ICDM_unsupervised_PCA},
and adopt the results of ConvAE from the original work \citep{Hasan.Mahmudul_etla_2016CVPR_DL_ConvAE}.
The results of ConvAE are already compared with recent state-of-the-art
baselines including supervised methods.

We follow similar procedures to what of our proposed framework for
OC-SVM and GMM, but apply these baselines on image patches clustered
by $k$-means. The kernel width and lower bound of the fraction of
support vectors of OC-SVM are set to $0.1$ and $10^{-4}$ respectively.
In GMM model, the number of Gaussian components is set to 20 and the
anomaly threshold is -50. These hyperparameters are also tuned to
obtain the best cross-validation results. It is noteworthy that it
is not straightforward to implement the incremental versions of the
baselines, thus we do not include them here.

\begin{figure}[h]
\centering{}\vspace{2mm}
\includegraphics[width=1\columnwidth]{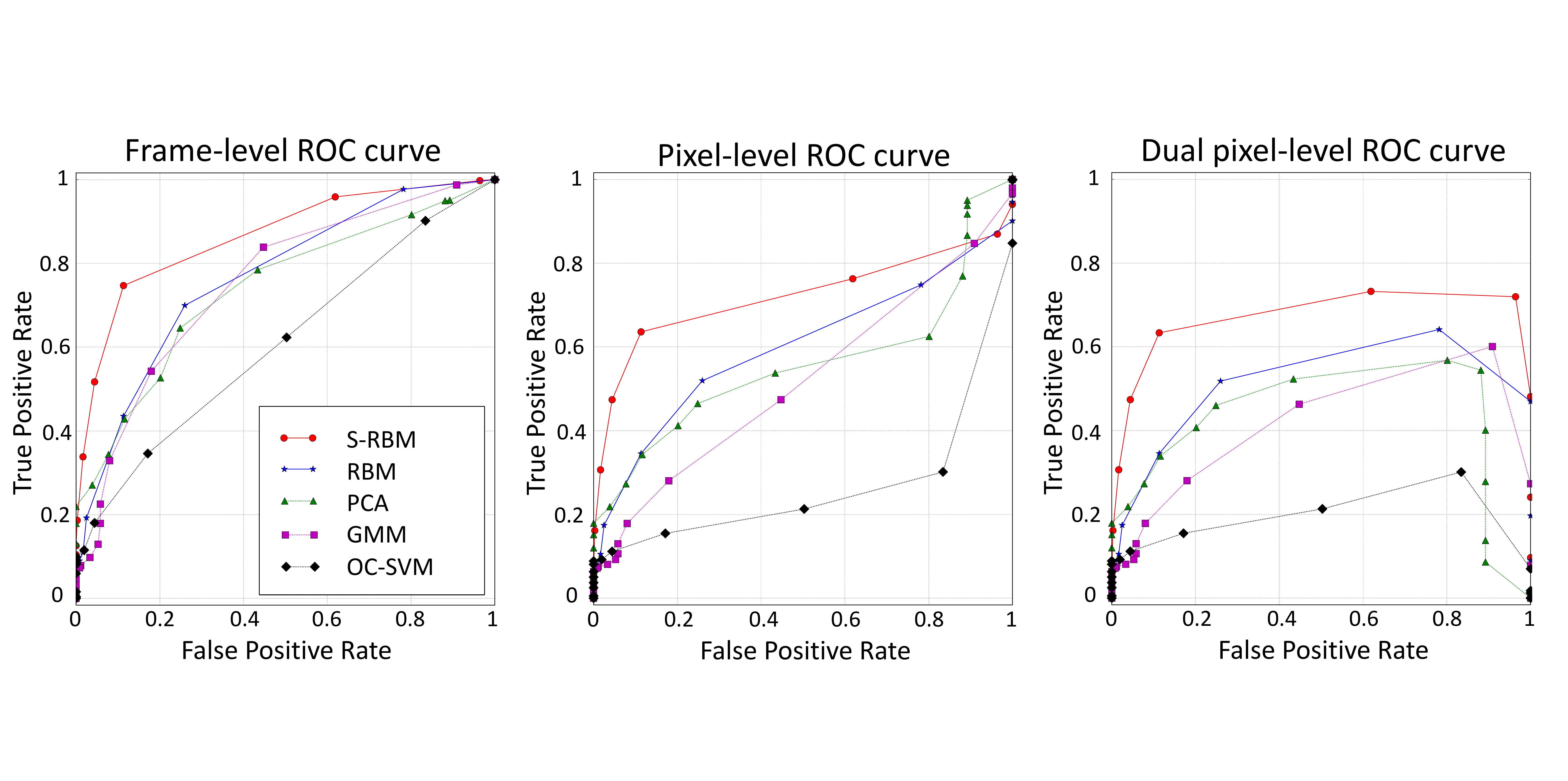}\vspace{-1mm}
\caption{Comparison ROC curves on UCSD Ped 2. Three figures share the same
legend. Higher curves indicate better performance. It is notable that,
unlike frame and pixel-level evaluations, dual-pixel level curves
may end at any points lower than (1,1). \label{figROC}}
\vspace{-1em}
\end{figure}

The ROC curves are shown in Fig.~\ref{figROC} whilst AUC and EER
scores are reported in Table~\ref{tableResult}. Both RBM and S-RBM
outperform the PCA, OC-SVM, GMM with higher AUC and lower EER scores.
Specially, our methods can produce higher AUC scores at dual pixel-level
which shows better quality in localizing anomaly regions. Additionally,
S-RBM achieves fairly comparable results with the ConvAE. It is noteworthy
that the ConvAE is a 12-layer deep architecture consisting of sophisticated
connections between its convolutional and pooling layers. On the other
hand, our $\model$ anomaly detector has only two layers, but obtains
a respectable performance. We believe that our proposed framework
is a promising system to detect abnormalities in video surveillance
applications. 

\definecolor{grey}{rgb}{0.9,0.9,0.9}

\begin{table}[h]
\begin{centering}
\vspace{3mm}
\resizebox{1.0\textwidth}{!}{
\begin{tabular}{>{\raggedleft}m{1.5cm}rr|rr|r|rr|rr|r|rr|rr|r}
\cline{2-16} 
\multirow{3}{1.5cm}{} & \multicolumn{5}{c|}{\textbf{Ped1}} & \multicolumn{5}{c|}{\textbf{Ped2}} & \multicolumn{5}{c}{\textbf{Avenue}}\tabularnewline
\cline{2-16} 
 & \multicolumn{2}{c|}{Frame} & \multicolumn{2}{c|}{Pixel} & Dual & \multicolumn{2}{c|}{Frame\textbf{ }} & \multicolumn{2}{c|}{Pixel} & Dual & \multicolumn{2}{c|}{Frame} & \multicolumn{2}{c|}{Pixel} & Dual\tabularnewline
 & AUC & EER & AUC & EER & AUC & AUC & EER & AUC & EER & AUC & AUC & EER & AUC & EER & AUC\tabularnewline
\hline 
PCA  & 60.28 & 43.18 & 25.39 & 39.56 & 8.76 & 73.98 & 29.20 & 55.83 & 24.88 & 44.24 & 74.64 & 30.04 & 52.90 & 37.73 & 43.74\tabularnewline
\rowcolor{grey}OC-SVM & 59.06 & 42.97 & 21.78 & 37.47 & 11.72 & 61.01 & 44.43 & 26.27 & 26.47 & 19.23 & 71.66 & 33.87 & 33.16 & 47.55 & 33.15\tabularnewline
GMM & 60.33 & 38.88  & 36.64  & 35.07 & 13.60 & 75.20 & 30.95 & 51.93 & 18.46 & 40.33 & 67.27 & 35.84 & 43.06 & 43.13 & 41.64\tabularnewline
\rowcolor{grey}ConvAE & \textbf{81.00} & \textbf{27.90} & \_ & \_ & \_ & \textbf{90.00} & 21.70 & \_ & \_ & \_ & 70.20 & \textbf{25.10} & \_ & \_ & \_\tabularnewline
\hline 
\textbf{\emph{RBM}} & 64.83 & 37.94 & 41.87 & 36.54 & 16.06 & 76.70 & 28.56 & 59.95 & 19.75 & 46.13 & 74.88 & 32.49 & 43.72 & 43.83 & 41.57\tabularnewline
\rowcolor{grey}\textbf{\emph{S-RBM}}\textbf{ } & 70.25 & 35.40 & \textbf{48.87} & \textbf{33.31} & \textbf{22.07} & 86.43 & \textbf{16.47} & \textbf{72.05} & \textbf{15.32} & \textbf{66.14} & \textbf{78.76} & 27.21 & \textbf{56.08} & \textbf{34.40} & \textbf{53.40}\tabularnewline
\hline 
\end{tabular}}
\par\end{centering}

\begin{centering}
\vspace{2mm}

\par\end{centering}

\caption{Anomaly detection results (AUC and EER) at frame-level, pixel-level
and dual pixel-level ($\alpha=5\%)$ on 3 datasets. Higher AUC and
lower EER indicate better performance. Meanwhile, high dual-pixel
values point out more accurate localization. We do not report EER
for dual-pixel level because this number do not always exist. Best
scores are in bold. Note that the frame-level results of ConvAE are
taken from \citep{Hasan.Mahmudul_etla_2016CVPR_DL_ConvAE}, but the
pixel-level and dual-pixel level results are not available.\label{tableResult}}
\end{table}